\documentclass{IEEEtran}

\usepackage{alltt}
\usepackage{caption}
\usepackage{paralist}
\usepackage{nicefrac}
\usepackage{booktabs}
\usepackage{algorithm}
\usepackage{algorithmic}
\usepackage{xspace}
\usepackage{amsmath}
\usepackage{svg}
\usepackage{marvosym}
\usepackage{wasysym}
\usepackage{verbatim}
\usepackage{subfigure}
\usepackage{hyperref}
\usepackage{multirow}
\usepackage{cite}
\usepackage[capitalize]{cleveref}
\usepackage[per-mode=symbol,detect-all]{siunitx}
\usepackage{graphics}
\usepackage{amssymb}
\usepackage{wrapfig}
\usepackage{enumitem}
\usepackage{placeins}

\usepackage{tikz,ifthen,pgfplots}

\usepackage{arydshln}
  
\usetikzlibrary{arrows,trees,backgrounds,automata,shapes,decorations,plotmarks,fit,calc,positioning,shadows,chains}

\definecolor{color0}{RGB}{228,87,46}
\definecolor{color1}{RGB}{23,190,187}
\definecolor{color2}{RGB}{255,201,20}
\definecolor{color3}{RGB}{46,40,42}
\definecolor{color4}{RGB}{118,176,65}

\definecolor{color0}{RGB}{250,121,33}
\definecolor{color1}{RGB}{254,153,32}
\definecolor{color2}{RGB}{185,164,76}
\definecolor{color3}{RGB}{86,110,61}
\definecolor{color4}{RGB}{12,71,103}

\definecolor{color0}{RGB}{228,253,225}
\definecolor{color1}{RGB}{138,203,136}
\definecolor{color2}{RGB}{100,131,129}
\definecolor{color3}{RGB}{87,87,97}
\definecolor{color4}{RGB}{255,191,70}

\definecolor{color0}{RGB}{226,59,62}
\definecolor{color1}{RGB}{243,114,44}
\definecolor{color2}{RGB}{248,150,30}
\definecolor{color3}{RGB}{249,199,79}
\definecolor{color4}{RGB}{126,179,86}
\definecolor{color5}{RGB}{67,170,139}
\definecolor{color6}{RGB}{39,125,161}
\definecolor{color7}{RGB}{21,49,60}
\definecolor{color8}{RGB}{180,215,228}

\tikzstyle{every pin edge}=[<-,shorten <=1pt]
\tikzstyle{neuron}=[circle,fill=black!25,minimum size=17pt,inner sep=0pt]
\tikzstyle{input neuron}=[neuron, fill=green!50]
\tikzstyle{output neuron}=[neuron, fill=red!50]
\tikzstyle{hidden neuron}=[neuron, fill=blue!50]
\tikzstyle{annot} = [text width=4em, text centered]
\usetikzlibrary{calc}

\newcommand{\mysubsubsection}[1]{\medskip\noindent\textbf{#1}}

\newcommand{\relu}{\text{ReLU}\xspace{}}
\newcommand{\elu}{\text{ELU}\xspace{}}

\newcommand{\sat}{\texttt{SAT}}
\newcommand{\unsat}{\texttt{UNSAT}}
\newcommand{\timeout}{\texttt{TIMEOUT}}

\newcommand{\ensemble}{\mathcal{E}}

\newcommand{\ME}{\mathrm{ME}}

\newcommand{\US}{\mathrm{US}}

\newcommand\blfootnote[1]{%
	\begingroup
	\renewcommand\thefootnote{}\footnote{#1}%
	\addtocounter{footnote}{-1}%
	\endgroup
}

\definecolor{nnedgecolor}{RGB}{90,90,90}
\tikzstyle{every pin edge}=[<-,shorten <=1pt]
\tikzstyle{every path}=[draw=color7!50]
\tikzstyle{neuron}=[circle,fill=black!25,minimum size=17pt,inner sep=0pt]
\tikzstyle{input neuron}=[neuron, fill=color4]
\tikzstyle{output neuron}=[neuron, fill=color0]
\tikzstyle{hidden neuron}=[neuron, fill=color6!80]
\tikzstyle{annot} = [text width=4em, text centered]
\tikzstyle{nnedge} = [-{stealth},shorten >=0.1cm, shorten <=0.05cm,line 
width=0.8pt,nnedgecolor]
\usetikzlibrary{calc}

\newtheorem{definition}{\textbf{Definition}}

\begin{document}

\title{Verification-Aided Deep Ensemble Selection}


\author{ Guy Amir, Tom Zelazny, Guy Katz and Michael Schapira\\ The Hebrew 
University of Jerusalem, Jerusalem, Israel \\ $ \lbrace $guyam, tomz, guykatz, schapiram$ 
\rbrace $@cs.huji.ac.il}

\maketitle

\begin{abstract}
  Deep neural networks (DNNs) have become the technology of choice for
  realizing a variety of complex tasks. However, as highlighted by
  many recent studies, even an imperceptible perturbation to a
  correctly classified input can lead to misclassification by a
  DNN. This renders DNNs vulnerable to strategic input manipulations
  by attackers, and also oversensitive to environmental
  noise. To mitigate this phenomenon, practitioners apply joint
  classification by an \textit{ensemble} of DNNs. By aggregating the
  classification outputs of different individual DNNs for the same
  input, ensemble-based classification reduces the risk of
  misclassifications due to the specific realization of the stochastic
  training process of any single DNN. However, the effectiveness of a
  DNN ensemble is highly dependent on its members \emph{not
    simultaneously erring} on many different inputs.  In this
  case study, we harness recent advances in DNN verification to devise
  a methodology for identifying ensemble compositions that are less
  prone to simultaneous errors, even when the input is adversarially
  perturbed --- resulting in more \textit{robustly-accurate} ensemble-based
  classification.  Our proposed framework uses a DNN verifier as a
  backend, and includes heuristics that help reduce the high
  complexity of directly verifying ensembles.  More broadly, our work
  puts forth a novel universal objective for formal verification that
  can potentially improve the robustness of real-world,
  deep-learning-based systems across a variety of application domains.
	\blfootnote{[*] This is an extended version
	of a paper with the same title that appeared at FMCAD 2022.}
\end{abstract}

\section{Introduction}
\label{sec:introduction}

In recent years, deep learning~\cite{GoBeCo16} has emerged as the
state-of-the-art solution for a myriad of tasks. Through the automated
training of \emph{deep neural networks} (\emph{DNNs}), engineers can
create systems capable of correctly handling previously unencountered
inputs. DNNs excel at tasks ranging from image
recognition and natural language processing to game playing and
protein folding~\cite{SiZi14, SiHuMaGuSiVaScAnPaLaDi16, Al19,
  CoWeBoKaKaKu11, HiDeYuDaMoJaSeVaNgSaKi12, KrSuHi12}, and are expected to play 
  a key role in various complex
systems~\cite{BoDeDwFiFlGoJaMoMuZhZhZhZi16,
  JuLoBrOwKo16}.

Despite their immense success, DNNs suffer from severe
vulnerabilities and weaknesses. A prominent example is the sensitivity
of DNNs to \emph{adversarial inputs}~\cite{SzZaSuBrErGoFe13, GoShSz14,
  KuGoBe16}, i.e., slight perturbations of correctly-classified inputs
that result in misclassifications. The susceptibility of DNNs to input
perturbations involves two risks that limit the applicability of deep
learning to mission-critical tasks: (1) falling victim to strategic
input manipulations by \textit{attackers}, and (2) failing to
\emph{generalize} well in the presence of environmental noise. In
light of the above, recent work has focused on enhancing the 
\textit{robustness} of DNN-based
classification to adversarial inputs while preserving 
\textit{accuracy}~\cite{GaUsAjGeLaLaMaLe16, TaKuPaGoBoMc17, ZhYuJiXiElJo19, 
MoYaCh21, BhJhCh21}.
 Informally,
a classifier is \textit{robustly accurate} (aka
\textit{astute}~\cite{WaJhCh18}) with respect to a given
distribution over inputs, if it continues to correctly classify inputs
drawn from this distribution, with high probability, even when these
inputs are arbitrarily perturbed (up to some maximally allowed perturbation).


We focus here on a classic technique for improving classification
quality~\cite{LePuCoCrBa15, ArCoSaIgCa17}: combining the outputs of an
\textit{ensemble}~\cite{HaSa90, Ta19, FoHuLa19} of DNN-based classifiers on an input to
derive a joint classification decision for that input.  By
incorporating the outputs of \textit{independently-trained} DNNs,
ensembles mitigate the risk of misclassification of a single DNN due
to a specific realization of its stochastic training process and the
specifics of its training data traversal. For a DNN ensemble to
provide a meaningful improvement over utilizing a single DNN, its
members should not frequently misclassify \textit{the same}
input. Consider, for instance, an extreme example, where an ensemble
with $k=10$ members is used, but for some part of the input space, the $10$
DNNs effectively behave identically, making mistakes on the exact same
inputs. In this scenario, the ensemble as a whole is no more robust on
this input subspace than each of its individual members. Our objective
is to demonstrate how recent advances in DNN
verification~\cite{KaBaDiJuKo17Reluplex, HuKwWaWu17} can be harnessed
to provide system designers and engineers with the means to avoid such
scenarios, by constructing adequately diverse ensembles.

Significant progress has recently been made on formal
verification techniques for DNNs~\cite{LuScHe21, AlAvHeLu20,
  AvBlChHeKoPr19, BaShShMeSa19, PrAf20, AnPaDiCh19, SiGePuVe19,
  XiTrJo18, Eh17}.  The basic DNN verification query is to
determine, given a DNN $N$, a precondition $P$, and a postcondition
$Q$, whether there exists an input $x$ such that $P(x)$ and $Q(N(x))$
both hold. Recent verification work has focused on \textit{identifying} adversarial inputs to DNN-based classification, or formally proving that no such inputs exist~\cite{GeMiDrTsChVe18, LyKoKoWoLiDa20, GoKaPaBa18}.  We demonstrate the applicability of DNN verification to solving a
new kind of queries, pertaining to DNN ensembles, which could
significantly boost the robustness of these ensembles (as opposed to
just measuring the robustness of individual DNNs).  We note that
despite great strides in recent years~\cite{SiGePuVe19,
  LyKoKoWoLiDa20, KaHuIbJuLaLiShThWuZe19}, even state-of-the-art DNN
verification tools face severe scalability limitations. This renders
solving verification queries pertaining to ensembles extremely
challenging, since the complexity of this task grows exponentially
with the number of ensemble members (see Section~\ref{sec:approach}).

In this case-study paper, we propose and evaluate an efficient
and scalable approach for verifying that different ensemble members do
not tend to err simultaneously. Specifically, our scheme considers
\textit{small subsets} of ensemble members,\footnote{While our technique is 
applicable to subsets of any size, we focused on pairs in our 
evaluation, as we later elaborate.} and dispatches
verification queries to seek perturbations of inputs for which \textit{all}
members in the subset err \textit{simultaneously}. By identifying such
inputs, we can assign a \textit{mutual error score} to each
subset. Using these mutual error scores, we compute, for each
individual ensemble member, a \textit{uniqueness score} that signifies how often it errs
simultaneously with other ensemble members. This score can be used to
detect the ``weakest'' ensemble members, i.e. those most prone to
erring in parallel to others, and replace them with fresh DNNs ---
thus enhancing the diversity among the ensemble members, and improving the overall robust accuracy of the ensemble.

To evaluate our scheme, we implemented it as a proof-of-concept tool,
and used this tool to conduct extensive experimentation on DNN
ensembles for classifying digits and clothing items. Our results
demonstrate that by identifying the weakest ensemble members (using
verification) and replacing them, the robust accuracy of the ensemble
as a whole may be significantly improved.  Additional experiments that
we conducted also demonstrate that our verification-driven approach
affords significant advantages when compared to competing,
non-verification-based, methods. Together, these results showcase the
potential of our approach. Our code and benchmarks are 
publicly available online~\cite{ArtifactRepository}.

The rest of the paper is organized as follows.
Section~\ref{sec:background} contains background on DNN ensembles and
DNN verification. In Section~\ref{sec:approach} we present our
verification-based methodology for ensemble selection, and then
present our case study in Section~\ref{sec:caseStudy}. Next, in
Section~\ref{sec:gradientAttacks} we compare our verification-based
approach to state-of-the-art, gradient-based, methods. Related work is
covered in Section~\ref{sec:relatedWork}, and we conclude and discuss
future work in Section~\ref{sec:conclusion}.

\section{Background}
\label{sec:background}

\mysubsubsection{Deep Neural Networks.}
A deep neural network (DNN)~\cite{GoBeCo16} is a directed graph,
comprised of layers of nodes (also known as \emph{neurons}). In
feed-forward DNNs, data flows sequentially from the first
(\emph{input}) layer, through a sequence of intermediate
(\emph{hidden}) layers, and finally into an \textit{output}
layer. The network's output is evaluated by assigning values to the
input layer's neurons and computing the value assignment for neurons
in each of the following layers,
in order, until reaching the output layer and returning its neuron
values to the user. In classification networks, which are our subject
matter here, each output neuron corresponds to an output \emph{class};
and the output neuron with the highest value represents the class, or
label, which the particular input is being classified as.

\begin{figure}[htp]
	\begin{center}
		\scalebox{1.0} {
			\def\layersep{2.0cm}
			\begin{tikzpicture}[shorten >=1pt,->,draw=black!50, node 
			distance=\layersep,font=\footnotesize]
				
				\node[input neuron] (I-1) at (0,-1) {$v^1_1$};
				\node[input neuron] (I-2) at (0,-2.5) {$v^2_1$};
				
				\node[hidden neuron] (H-1) at (\layersep,-1) {$v^1_2$};
				\node[hidden neuron] (H-2) at (\layersep,-2.5) {$v^2_2$};
				
				\node[hidden neuron] (H-3) at (2*\layersep,-1) {$v^1_3$};
				\node[hidden neuron] (H-4) at (2*\layersep,-2.5) {$v^2_3$};
				
				\node[output neuron] at (3*\layersep, -1) (O-1) {$v^1_4$};
				\node[output neuron] at (3*\layersep, -2.5) (O-2) {$v^2_4$};
				
				\draw[nnedge] (I-1) --node[above] {$1$} (H-1);
				\draw[nnedge] (I-1) --node[above, pos=0.3] {$\ -3$} (H-2);
				\draw[nnedge] (I-2) --node[below, pos=0.3] {$2$} (H-1);
				\draw[nnedge] (I-2) --node[below] {$-1$} (H-2);
				
				\draw[nnedge] (H-1) --node[above] {$\relu$} (H-3);
				\draw[nnedge] (H-2) --node[below] {$\relu$} (H-4);
				
				\draw[nnedge] (H-3) --node[above] {$4$} (O-1);
				\draw[nnedge] (H-4) --node[below, pos=0.3] {$6$} (O-1);
				\draw[nnedge] (H-3) --node[above, pos=0.3] {$-1$} (O-2);
				\draw[nnedge] (H-4) --node[below] {$3$} (O-2);

				\node[below=0.05cm of H-1] (b1) {$+1$};
				\node[below=0.05cm of H-2] (b2) {$-1$};
				
				\node[annot,above of=H-1, node distance=0.8cm] (hl1) {Weighted 
				sum};
				\node[annot,above of=H-3, node distance=0.8cm] (hl2) {ReLU };
				\node[annot,left of=hl1] {Input };
				\node[annot,right of=hl2] {Output };
			\end{tikzpicture}
		}
		\captionsetup{size=small}
		\captionof{figure}{A toy DNN.}
		\label{fig:toyDnn}
	\end{center}
\end{figure}
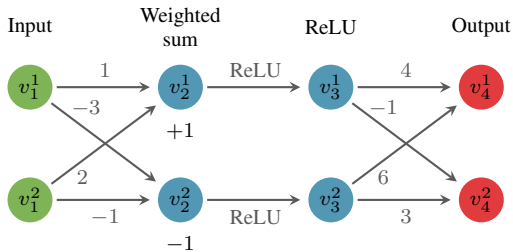
Fig.~\ref{fig:toyDnn} depicts a toy DNN. It has an input layer with
two neurons, followed by a \emph{weighted sum layer}, which computes
an affine transformation of values from its preceding layer. For
example, for input $V_1=[1, -5]^T$, the second layer's computed values
are $V_2=[-8,1]^T$. Next is a \relu{} layer, which applies the \relu{}
function $\relu{}(x)=\max(0,x)$ to each individual neuron, resulting
in $V_3=[0,1]^T$. Finally, the network's output layer again computes
an affine transformation, resulting in the output $V_4=[6,3]^T$. Thus,
input $[1,-5]^T$ is classified as the label corresponding to neuron
$v^1_4$. For additional details, see~\cite{GoBeCo16}.

\mysubsubsection{Accuracy, Robustness, and Deep Ensembles.}
The weights of a DNN are determined through its training process. In
supervised learning, we are provided a set of pairs $(x_i,l_i)$ drawn according to some
(unknown) distribution $D$, where
$x_i$ is an input point and $l_i$ is a ground-truth label for that
input. The goal is to select weights for the DNN
$N$ that maximize its \textit{accuracy}, which is defined as:
$Pr_{(x,l)\sim D}(N(x)=l)$ (we slightly abuse notation, and use
$N(x)$ to denote both the network's output vector, as well as the
label it assigns $x$).

\begin{figure}[htp]
  \begin{center}
   \includegraphics[width=0.7\linewidth]{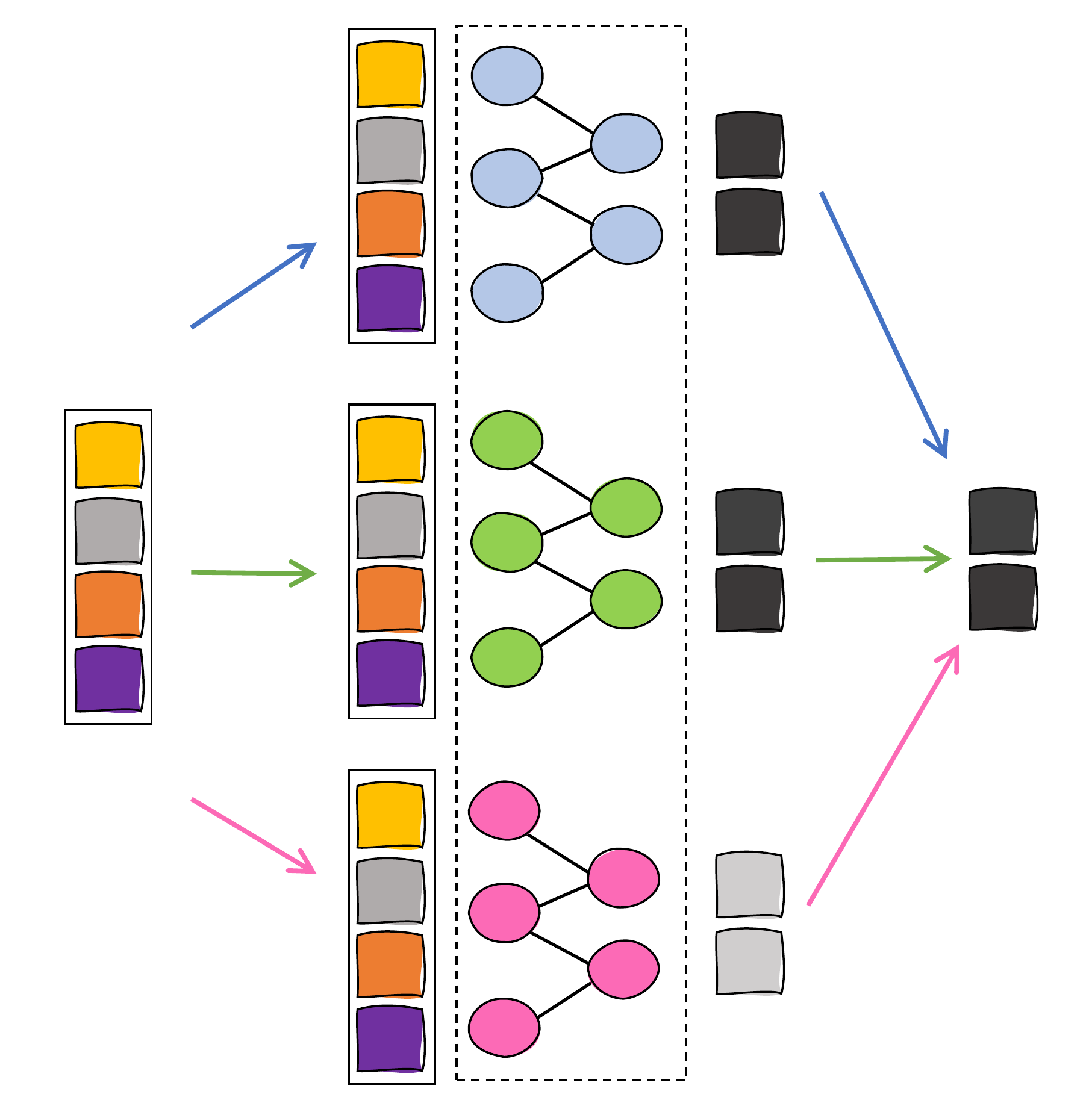}
  \end{center}
  \caption{An ensemble
    comprising three DNNs. Each input vector is
    independently classified by all three networks, and the results
    are aggregated into a final classification. }
	\label{fig:toyEnsemble}
\end{figure}

We restrict our attention to the \textit{classification} setting, in
which labels are discrete.
The training of a DNN-based classifier is typically a
stochastic process. This process is affected, for example, by the initial assignment of
weights to the DNN, the order in which training data is traversed, and
more. A prominent method for avoiding misclassifications originating from the stochastic training of a single DNN is
employing \textit{deep ensembles}. A deep ensemble is a set
$\ensemble =\{N_1,\ldots, N_k\}$ of $k$ independently-trained
DNNs. The ensemble classifies an
input by aggregating the individual classification outputs of its
members (see Fig.~\ref{fig:toyEnsemble}). The collective decision is
typically achieved by averaging over all members' outputs.
Ensembles
have been shown to often achieve better accuracy than their individual
members~\cite{LePuCoCrBa15,ArCoSaIgCa17, XuSoPl18,
  LyGuRaBe20}.

A critical condition for the success of ensemble-based classifiers is
that the ensemble members' misclassifications are not strongly 
correlated~\cite{SvBi07, NaYoLeLe21, 
LePuCoRaCrBa16}. This key property is crucial in order to avoid a scenario 
  where
many different members of the ensemble frequently make mistakes on the
same input, causing the ensemble as a whole to also err on that
input. Heuristics for achieving diversity across ensemble members
include, e.g., training the members simultaneously with
diversity-aware loss~\cite{LePuCoCrBa15, JaLuMuGi20}, randomly
initializing different weights for the ensemble
members~\cite{LaPrBl16}, and other methods~\cite{ShMoMaHeGa18, NaYoLeLe21}.

Since the discovery of adversarial inputs, practitioners have become
interested in DNNs that are not only accurate but also \emph{robustly
  accurate}.  We say that a network $N$ is $\epsilon$-robust around
the point $x$ if every input point that is at most $\epsilon$ away
from $x$ receives the same classification as $x$:
$\lVert x'-x\rVert\leq \epsilon \Rightarrow N(x)=N(x')$, where $N(x)$
is the label assigned to $x$; and the definition of accuracy is
generalized to $\epsilon$-robust accuracy as follows:
$Pr_{(x,l)\sim D}(\lVert x'-x\rVert\leq \epsilon \Rightarrow N(x')=l).
$
While improvements in accuracy afforded by ensembles are
straightforward to measure, this is typically not the case for robust
accuracy, as we discuss in Section~\ref{sec:approach}.

\mysubsubsection{DNN Verification.}
Given a DNN $N$, a verification query on $N$ specifies a precondition
$P$ on $N$'s input vector $x$, and a postcondition $Q$ on $N$'s output
vector $N(x)$. A DNN verifier needs to determine whether there exists
a concrete input $x_0$ that satisfies $P(x_0)\wedge Q(N(x_0))$ (the
\sat{} case), or not (the \unsat{} case). Typically, $P$ and $Q$ are
expressed in the logic of linear real arithmetic.  For instance, the
$\epsilon$-robustness of a DNN around a point $x$ can be phrased as a
DNN verification query, and then dispatched using existing
technology~\cite{GeMiDrTsChVe18, WaPeWhYaJa18, KaBaDiJuKo17Reluplex}.
The DNN verification problem is known to be
NP-complete~\cite{KaBaDiJuKo21}.


\section{Improving Robust Accuracy using Verification}
\label{sec:approach}

\subsection{Directly Quantifying Robust Accuracy is Hard}\label{subsec:directly}

In order to construct a robustly-accurate ensemble $\ensemble$ with
$k$ members, we train a set of $n>k$ DNNs and then seek to select a
subset of $k$ DNNs that provides high robust accuracy.  This method of
training multiple models and then discarding a subset thereof is known
as \emph{ensemble pruning}, and is a common practice in deep-ensemble
training~\cite{ZhWuTa02, BiWaYiYaCh19}. In our case, a 
straightforward approach to do so would be to quantify the robust
accuracy for all possible $k$-sized DNN-subsets, and then pick the
best one. This, however, is computationally expensive, and requires an
accurate estimate of the robust accuracy of an ensemble.

A natural approach for estimating the $\epsilon$-robust accuracy of a
DNN is to verify, for many points in the test data,
that the DNN yields an accurate label not only on each data
point itself, but also on each and every input derived from that data
point via an $\epsilon$-perturbation~\cite{GeMiDrTsChVe18}. The
fraction of tested points for which this is
indeed the case can be used to estimate the accuracy of the classifier
on the underlying distribution from which the data is
generated.

A similar process can be performed for an ensemble
$\ensemble = \{N_1,\ldots,N_k\}$, by first constructing a single,
large DNN $N_\ensemble$ that aggregates $\ensemble$'s joint
classification, and then verifying its robustness on a set of points
from the test data (see
Section~\ref{sec:appendix:verifyingAnEnsemble} of the
Appendix). However, this approach faces a significant scalability
barrier: the DNN ensemble, $N_\ensemble$, comprised of all $k$ member-DNNs is 
(roughly) 
$k$ times larger than any of the
$N_i$'s, and since DNN verification becomes exponentially harder as
the DNN size increases, $N_\ensemble$'s size might render efficient
verification infeasible. As we demonstrate later, this is the case
even when the constituent networks themselves are fairly small.  Our
proposed methodology circumvents this difficulty by only solving
verification queries pertaining to \textit{very small} sets of DNNs.
\subsection{Mutual Error Scores and Uniqueness Scores}


In general, the less likely it is that members of an ensemble err
simultaneously with other members, the more accurate the ensemble
is. This motivates our definition of mutual error scores below.


\indent
\begin{definition}[\textbf{Agreement Points}]
  Given an ensemble $\ensemble = \{N_1,N_2,\ldots,N_k\}$, we say that
  an input point $x_0$ is an \emph{agreement point} for $\ensemble$ if
  there is some label $y_0$ such that $N_i(x_0)=y_0$ for all
  $i\in [k]$. We let $\ensemble(x_0)$ denote the label $y_0$.
\end{definition}

As we later discuss, the $\epsilon$-neighborhoods of agreement points
are natural locations for detecting hidden tendencies of 
ensemble members to err together.

\indent
\begin{definition}[\textbf{Mutual Errors}]
  Let $\ensemble$ be an ensemble, and let $x_0$ be an agreement point for $\ensemble$. Let $B_{x_0,\epsilon}$ be the $\epsilon$-ball around $x_0$,
  $
    B_{x_0,\epsilon} = \{ x \ |\ \lVert x - x_0\rVert_{\infty}\leq
    \epsilon\}.
  $
  We say that $N_1$ and $N_2$ have a \textit{mutual error} in $B$ if there exists a point $x\in B_{x_0,\epsilon}$ such that $N_1(x)\neq \ensemble(x_0)$
  and $N_2(x)\neq \ensemble(x_0)$.
\end{definition}

\indent

Intuitively, if $N_1$ and $N_2$ have many mutual errors, incorporating both into an ensemble is a poor choice. 
This naturally gives rise to the following definition:

\indent
\begin{definition}[\textbf{Mutual Error Scores}]
  Let $A$ be a finite set of $m$ agreement points in an ensemble $\ensemble$'s input
  space, and let $B_1,B_2,\ldots,B_m$ denote the $\epsilon$-balls
  surrounding the points in $A$. Let $N_1$, $N_2$ denote two members
  of $\ensemble$. The \emph{mutual error score} of $N_1$ and $N_2$ with
  respect to $\ensemble$ and $A$ is denoted by $\ME_{\ensemble,A}(N_1,N_2)$, 
  and defined as:
  \begin{align*}
    \ME_{\ensemble,A}&(N_1,N_2) = \\
    &\frac{| \{ i \ |\ N_1 \text{ and } N_2 \text{\ have a mutual error
        in\ } B_i\} | }{m}
  \end{align*}
\end{definition}
Observe that $\ME_{\ensemble,A}(N_1,N_2)$ is always in the range
$[0,1]$. The closer it is to $1$, the
more mutual errors $N_1$ and $N_2$ have, making it unwise to place
them in the same ensemble.

\indent
\begin{definition}[\textbf{Uniqueness Scores}]
  Given an ensemble $\ensemble = \{N_1,N_2,\ldots,N_n\}$ and a set $A$
  of agreement points for $\ensemble$, we define, for each ensemble member
  $N_i$, the \emph{uniqueness score} for $N_i$ with respect to
  $\ensemble$ and $A$, $\US_{\ensemble,A}(N_i)$, as:
  \[
    \US_{\ensemble,A}(N_i) = 1 - \frac{\sum_{j\neq i}\ME_{\ensemble,A}(N_i,N_j)}{n-1}
    \]
\end{definition}

The uniqueness score ($\US$) of $N_i$ is the complement of its
average mutual error score with the other ensemble members.  When this
score is close to $0$, $N_i$ tends to err simultaneously with other
members of the ensemble on points in $A$. In contrast, the closer the
uniqueness score is to $1$, the rarer it is for $N_i$ to misclassify
the same inputs as other members of the ensemble. Hence, ensemble
members with low uniqueness scores are, intuitively, good
candidates for replacement.

We point out that our definitions above can naturally be generalized to
larger subsets of the ensemble members --- thus measuring robust accuracy
more precisely, but rendering these measurements more
complex to perform in practice.


\mysubsubsection{Computing Mutual Errors.}  The only computationally
complex step in determining the uniqueness scores of individual
ensemble members is computing the pairwise mutual errors for the
ensemble. To this end, we leverage DNN verification technology.
Specifically, given two ensemble members $N_1$ and $N_2$, an agreement
point $a$ for the ensemble with label $l$, and $\epsilon>0$, an
appropriate DNN verification query can be formulated as
follows. First, we construct from $N_1$ and $N_2$ a single, larger DNN
$N$, which captures $N_1$ and $N_2$ simultaneously processing a shared
input vector, side-by-side. This network $N$ is then passed to a DNN
verifier, with the precondition that the input be restricted to $B$,
an $\epsilon$-ball around $a$, and the postcondition that (1) among
$N$'s output neurons that correspond to the outputs of $N_1$, the
neuron representing $l$ not be maximal, and (2) among $N$'s output
neurons that correspond to the outputs of $N_2$, the neuron
representing $l$ not be maximal. Such queries are supported by most
available DNN verification engines. We note that this encoding (depicted in 
Figure~\ref{fig:mutualError}), where
two networks and their output constraints are combined into a single
query, is crucial for finding inputs on which both DNNs err
\textit{simultaneously}.  For additional details, see
Section~\ref{sec:appendix:mutualErrors} of the Appendix.
 

\subsection{Ensemble Selection using Uniqueness Scores}\label{subsec_choosing_ensembles}

\medskip
\noindent{\textbf{An Iterative Scheme.}}
Building on our verification-based method for computing mutual error
scores, we propose an iterative scheme for constructing an ensemble.
Our scheme consists of the following steps:
 \begin{enumerate}
\item independently train a set $\mathcal{N}$ of $n$ DNNs, and identify a set $A$ of $m$ agreement points
  that are \textit{correctly classified} by all $n$ DNNs.\footnote{In our experiments, we arbitrarily chose $k=5$, $n=10$ and 
  $m=200$.} This is done by sequentially checking points from the validation
  dataset;
\item arbitrarily choose an initial candidate ensemble $\ensemble$ of
  size $k<n$; 
 \item compute (using a verification engine backend) all
  mutual error scores for the DNN members comprising $\ensemble$, with respect to $A$;
\item compute the uniqueness score for each ensemble member, and
  identify a DNN member $N_l$ with a low score;
\item identify a fresh DNN $N_f$, not currently in $\ensemble$, that
  has a higher uniqueness score than $N_l$, if one exists, and
  replace $N_l$ with $N_f$.  Specifically, identify a DNN
  $N_f\in \mathcal{N} \setminus \ensemble$, such that the uniqueness
  score of $N_f$ with respect to the ensemble
  $\ensemble \setminus \{N_l\} \cup \{N_f\}$ and the point set $A$,
  namely $\US_{\ensemble \setminus \{N_l\} \cup \{N_f\},A}(N_f)$, is
  maximal. If this score is greater than $\US_{\ensemble,A}(N_l)$,
  replace $N_l$ with $N_f$, i.e. set
  $\ensemble := \ensemble \setminus \{N_l\} \cup \{N_f\}$; and
\item repeat Steps (3) through (5), until no $N_f$ is found or until
  the user-provided timeout or maximal iteration count are exceeded.
\end{enumerate}

Intuitively, after starting with an arbitrary ensemble, we run
multiple iterations, each time trying to improve the
ensemble. Specifically, we identify the ``weakest'' member of the
current ensemble, and replace it with a fresh DNN that obtains a
higher uniqueness score relevant to the remaining members --- thus
ensuring that each change that we make improves the overall robust
accuracy on the fixed set of agreement points.  

The greedy
search procedure is repeated for the new candidate ensemble, and so
on. The process terminates after a predefined number of iterations is
reached, when the process converges (no further improvement is
achievable on the fixed set of agreement points), or when a predefined timeout value is exceeded.

\noindent{\textbf{On the Importance of Agreement Points.}}
Our iterative scheme for constructing an ensemble starts with an
arbitrary selection of $k$ candidate members, and then computes the
uniqueness score for each member.  
As mentioned, the uniqueness scores are computed
with respect to a fixed set of agreement points, pre-selected from the
validation data (which is labeled data, not used for training the
DNNs).
  

We point out that agreement points are data points on which there is
overwhelming consensus among ensemble members, despite the specific
realization of the training process of each member. As such, agreement
points correspond to data points that are ``easy'' to label correctly.
Consequently, data points in close proximity of
an agreement point are rarely classified differently than the
agreement point by an individual ensemble member, let alone by
multiple members simultaneously. As our objective is to expose
implicit tendencies of ensemble members to err together, the close
neighborhood of agreement points is a natural area for seeking joint
deviations from the consensual label (which are expected to be
extremely rare). In our evaluation, we computed uniqueness scores
based solely on \textit{correctly-classified} agreement points and
ignored any incorrectly-classified agreement points.\footnote{For
  example, in our MNIST experiments $99.7\%$ of the agreement points
  were correctly classified by all individual DNNs, and by the
  ensemble as a whole.}

As we later demonstrate, a small set of correctly-classified agreement
points from the validation set can be used, in practice, to identify
ensemble members that tend to err simultaneously on \emph{other} data
points. We note that this is also the case even when the chosen agreement 
points are all identically labeled.

\medskip
\noindent{\textbf{Monotonicity and Convergence.}} 
Using our approach, an ensemble member is replaced with a fresh DNN
only if this replacement leads to \textit{strictly} fewer joint errors
with the \textit{remaining members} on the fixed set of
agreement points. Thus, the total number of joint errors decreases with
every replacement; and, as this number is
trivially lower-bounded by $0$, this (``potential-function'' style)
argument establishes the process's monotonicity and convergence.


By iteratively reducing the number of joint errors across all pairs of
chosen ensemble members, our iterative process improves the robust
accuracy of the resulting ensemble on the fixed set of agreement
points. This, however, does not guarantee improved robust accuracy
over the entire input domain. Nonetheless, we show in
Section~\ref{sec:caseStudy} that such an improvement does typically
occur in practice, even on randomly sampled subsets of input points
(which are not necessarily agreement points).

\section{Case Study: MNIST and Fashion-MNIST}
\label{sec:caseStudy}


Below, we present the evaluation of our methodology on two datasets: the MNIST dataset for handwritten digit recognition~\cite{Le98}, and the Fashion-MNIST dataset for clothing classification~\cite{XiRaVo17}. Our results for both datasets demonstrate that our technique facilitates choosing ensembles that provide high robust accuracy via relatively few, efficient verification queries. 

The considered datasets are conducive for our purposes since they
allow attaining high accuracy using fairly small DNNs, which enables
us to \textit{directly quantify} the robust accuracy of an entire
ensemble, by dispatching verification queries that would otherwise be
intractable (see Section~\ref{subsec:directly}). This provides the
ground truth required for assessing the benefits of our approach.  The
scalability afforded by our approach is crucial even for handling the
relatively modest-sized DNNs considered: on the MNIST data, for
instance, mutual-error verification queries for two ensemble members
typically took a few seconds, whereas verification queries involving
the full ensemble of five networks often timed out ($35\%$ of the
queries on the MNIST data timed out after $24$ hours, versus only
roughly $1\%$ of the pairwise mutual-error queries).  As constituent
DNN sizes and ensemble sizes increase, this gap in scalability is
expected to become even more significant.

Our verification queries were dispatched using the Marabou
verification engine~\cite{KaHuIbJuLaLiShThWuZe19} (although other
engines could also be used). Additional details regarding the encoding
of the verification queries appear in
Sections~\ref{sec:appendix:verifyingAnEnsemble}
and~\ref{sec:appendix:mutualErrors} of the Appendix, and additional
details about the experimental results appear in
Section~\ref{sec:appendix:additionalEvaluationDetails} therein.  We have 
publicly released our code, as well as all benchmarks and experimental
data, within an artifact accompanying this paper~\cite{ArtifactRepository}.

%

\mysubsubsection{MNIST.}  For this part of our evaluation, we trained
$10$ independent DNNs $\{N_1,\ldots,N_{10}\}$ over the MNIST
dataset~\cite{Le98}, which includes 28$\times$28 grayscale images of
$10$ handwritten digits (from ``0'' to ``9'').  Each of these networks
had the same architecture: an input layer of $784$ neurons, followed
by a fully-connected layer with $30$ neurons, a \relu{} layer, another
fully-connected layer with $10$ neurons, and a final softmax layer
with 10 output neurons, corresponding to the 10 possible digit
labels.\footnote{Although the DNNs all have the same size and 
architecture, common ensemble training processes randomly initialize their 
weights, and also randomly pick samples from the same training set 
(see~\cite{LaPrBl16}). This is the cause for diversity among ensemble members, 
which our algorithm later detects.} All networks achieved high accuracy rates of
$96.29\% - 96.57\% $ (see Table~\ref{table:DigitMnistUniqueness}).

\begin{table*}[ht]
  \centering
  \caption{Accuracy and uniqueness scores for the MNIST
    networks.  Uniqueness scores are measured with respect to the
  ensemble (either $\ensemble_1$ or $\ensemble_2$).}
  \scalebox{1.0}{
    \begin{tabular}{l|ccccccccccccc|}
        &
  \multicolumn{5}{c}{$\ensemble_1$} &&&&
  \multicolumn{5}{c|}{$\ensemble_2$} 
  \\
      \cline{2-6} 
      \cline{10-14}
        &
  $N_1$ &
  $N_2$ &
  $N_3$ &
  $N_4$ &
  $N_5$ &
        &&&
  $N_6$ &
  $N_7$ &
  $N_8$ &
  $N_9$ &
  $N_{10}$
  \\
  \midrule
  Accuracy &
  96.42\% &
  96.55\% &
  96.40\% &
  96.46\% &
  96.29\% &
      &&&
  96.44\% &
  96.48\% &
  96.57\% &
  96.51\% &
  96.46\% 
  \\
  $\US$ &
  90.75\% &
  \textbf{{\color{red}88.38\%}} &
  90.63\% &
  92.13\% &
  \textbf{{\color{red}88.63\%}} &
      &&&
  97.38\% &
  96.75\% &
  97.5\% &
  \textbf{{\color{green}98.88\%}} &
  \textbf{{\color{green}97.75\%}} 
 \end{tabular}
}
\label{table:DigitMnistUniqueness}
\end{table*}




After training, we arbitrarily constructed two distinct ensembles with
five DNN members each: $\ensemble_1=\{N_1,\ldots,N_5\}$ and
$\ensemble_2=\{N_6,\ldots,N_{10}\}$, with an accuracy of $97.8\%$ and
 $97.3\%$, respectively. Notice that the ensembles achieve a
higher accuracy over the test set than their individual members.

\begin{figure} [h]
	\centering
	{\includegraphics[width=1.0\linewidth]{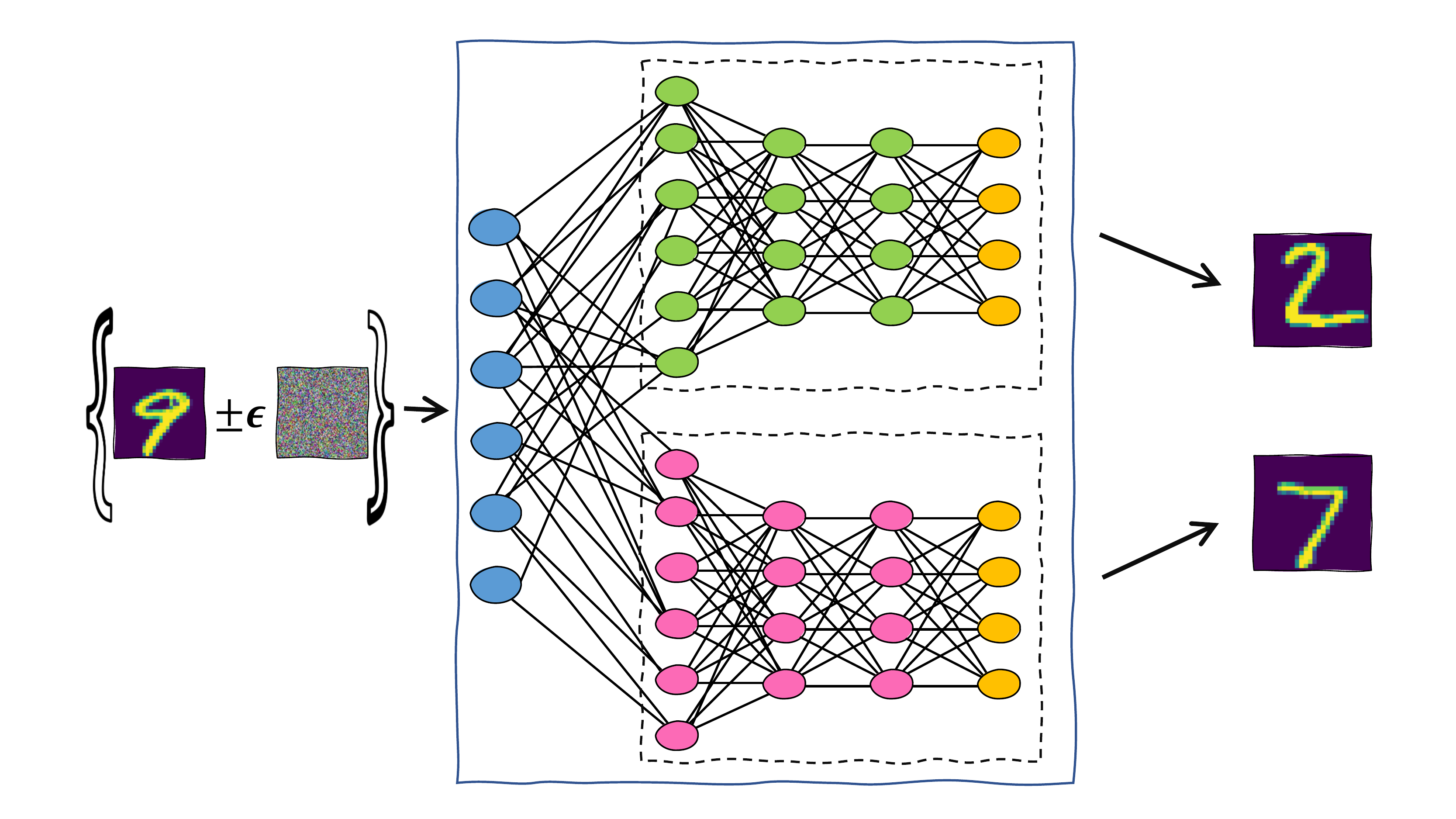}}%
	\caption{Checking whether two MNIST digit recognition networks
          have a mutual error around an agreement point labeled ``9''. In this 
          case, the same perturbation causes one network to
          output the incorrect label ``2'', and the other network to
          output the incorrect label ``7''.}%
	\label{fig:mutualError}%
\end{figure}

We then applied our method in an attempt to improve the
robust accuracy of $\ensemble_1$. We began by searching the validation
set, and identifying $200$ agreement points (the set $A$),\footnote{In our 
experiments, we empirically selected 200 agreement points in order to balance
 between precision (a higher number of points) and verification 
speed (a smaller number of points). This selection is based on a user’s 
available computing power.} all
correctly labeled as ``0'' by all 10 networks.\footnote{The “0” label is the 
label with the highest accuracy among the trained ensemble members, and thus 
``0''-labeled agreement points represent areas in the input space with 
extremely high consensus.} Using the $200$ agreement
points and 6 different perturbation sizes\footnote{$\epsilon$ values which are 
too small, or too
  large, render the queries trivial. Thus, we found
  it to be useful to use a varied selection of $\epsilon$ values.}
$\epsilon\in \{0.01, 0.02, 0.03, 0.04, 0.05, 0.06\}$, we
constructed $1200$ $\epsilon$-balls around the selected agreement points; and then, for
every ball $B$ and for every pair $N_i,N_j\in \ensemble_1$, we
encoded a verification query to check whether $N_i$ and $N_j$ have
a mutual error in $B$ (see example in  Fig.~\ref{fig:mutualError}). This resulted in
${5 \choose 2} \cdot 200 \cdot 6 = 12000$ verification queries, which
we dispatched using the Marabou DNN
verifier~\cite{KaHuIbJuLaLiShThWuZe19} (each query ran with a
2-hour timeout limit). Finally, we used the results to compute the uniqueness
score for each network in $\ensemble_1$; these results, which
appear briefly in Table~\ref{table:DigitMnistUniqueness} (for $\epsilon=0.02$) and 
appear in full in Table~\ref{table:DigitMnistEnsembleOneUniquenessScores} of
the Appendix, clearly show that two of the members, $N_2$ and $N_5$,
are each relatively prone to erring simultaneously with the remaining four members of $\ensemble_1$.

Next, we began searching among the remaining networks, $N_6,\ldots,N_{10}$,
for good replacements for $N_2$ and $N_5$. Specifically, we searched for
networks that obtained higher $\US$ scores than $N_2$ and $N_5$. To
achieve this, we began modifying $\ensemble_1$, each time removing
either $N_2$ or $N_5$, replacing them with one of the remaining
networks, and computing the uniqueness scores for the new members
(with respect to the four remaining original networks). We observed
that for both $N_2$ and $N_5$, network $N_9$ was a good replacement,
obtaining very high $\US$ values. For additional details, see
Section~\ref{sec:appendix:additionalEvaluationDetails} of the
Appendix.

\begin{figure*}[ht] 
    \centering
    \begin{minipage}{0.45\textwidth}
        \centering
        \includegraphics[width=1.1\textwidth]{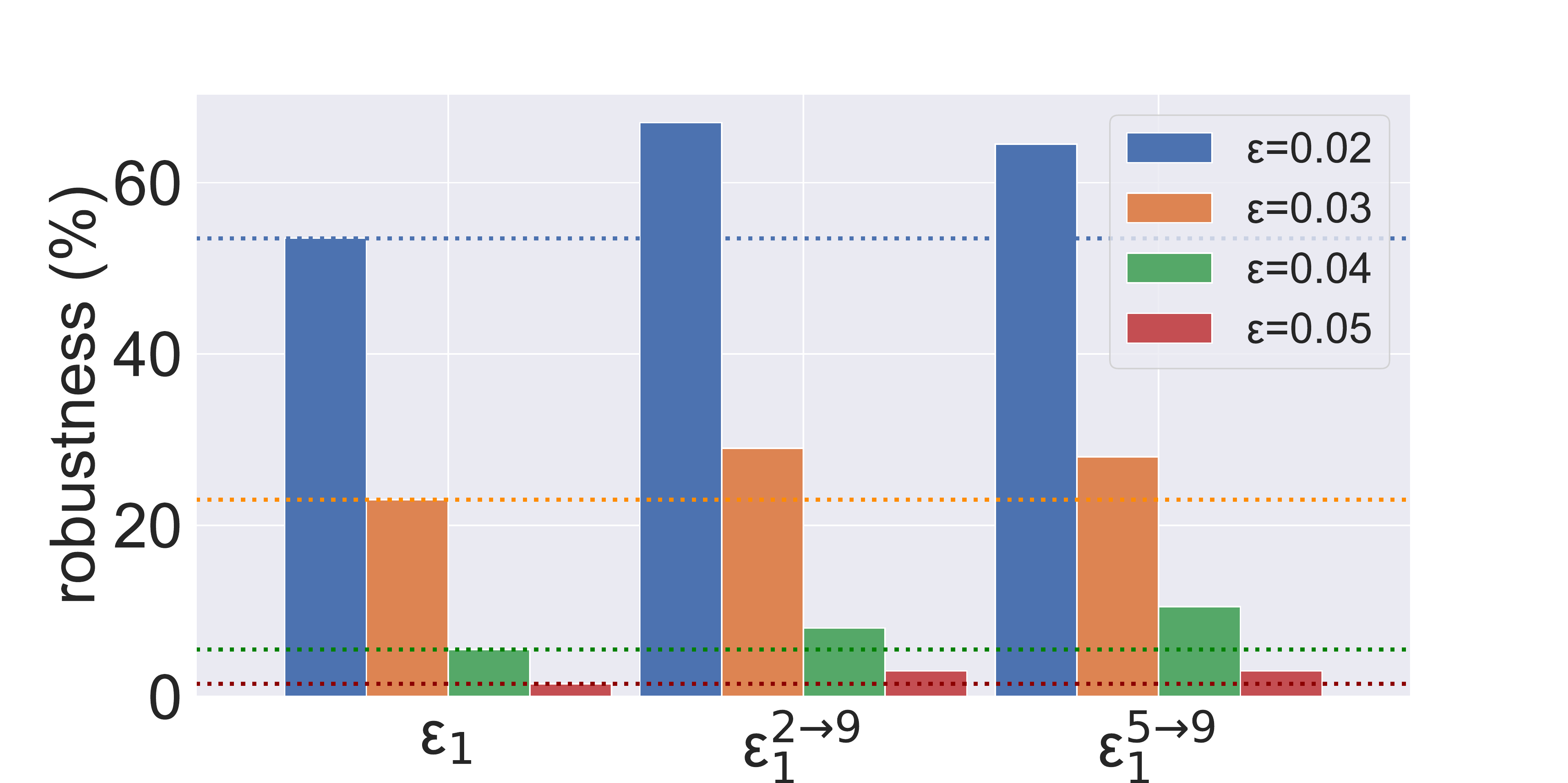}

    \end{minipage}\hfill
    \begin{minipage}{0.45\textwidth}
        \centering
        \includegraphics[width=1.1\textwidth]{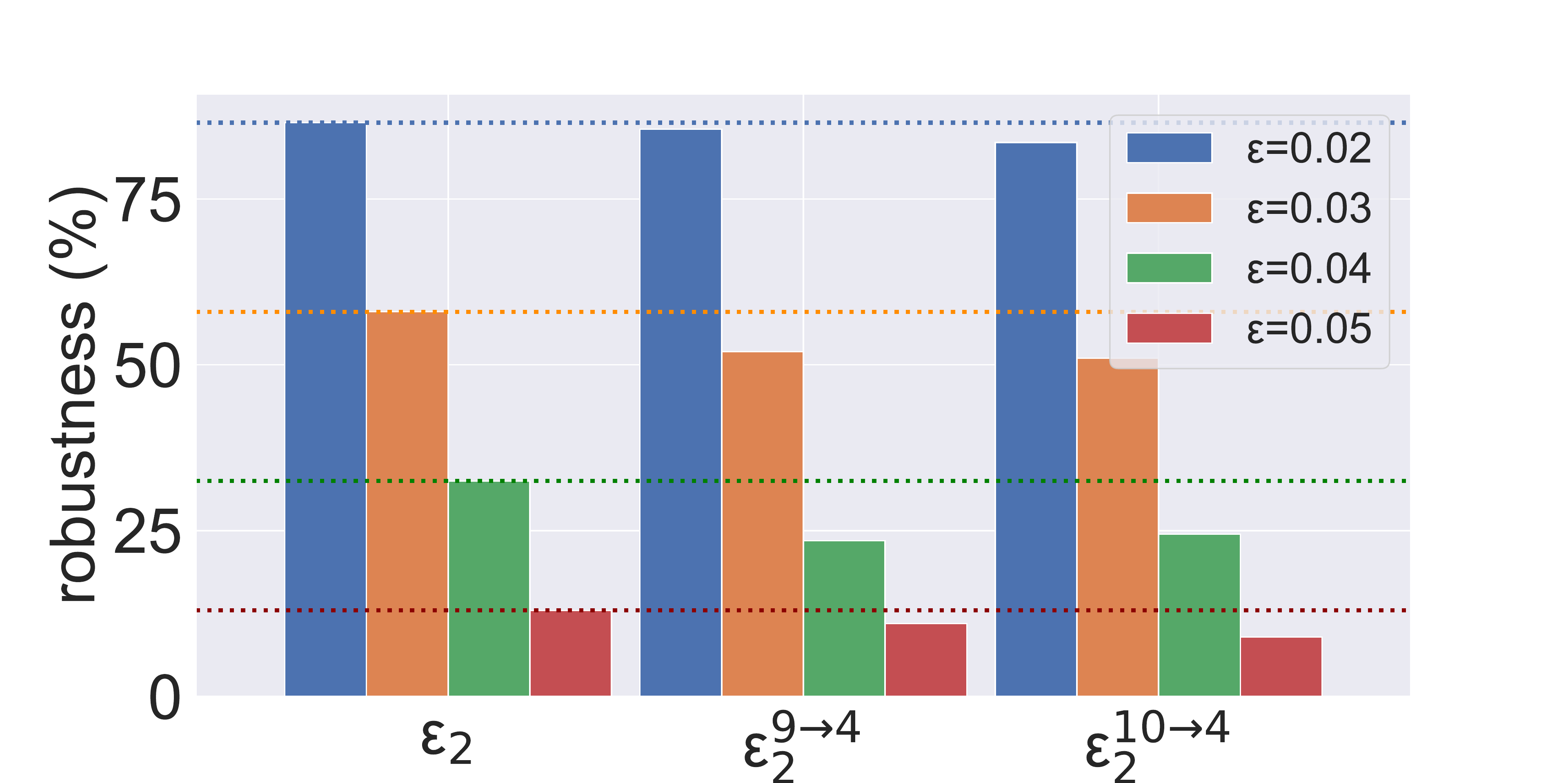}

            \end{minipage}
            \caption{The average robust accuracy scores for our
              original and modified ensembles.  The results for
              $\epsilon=0.01$ and $\epsilon=0.06$ are trivial (the
              ensembles achieve near-perfect or near-zero robustness),
              and are omitted to reduce clutter.}
            \label{fig:DigitMnistResultsRandomBatchEnsemble}
\end{figure*}

Finally, to evaluate the effect of our changes to $\ensemble_1$, we
constructed the two new ensembles,
$\ensemble_1^{2\rightarrow 9}=\{N_1,N_9,N_3,N_4,N_5\}$ and
$\ensemble_1^{5\rightarrow 9}=\{N_1,N_2,N_3,N_4,N_9\}$.  
Computing the new ensembles' robust accuracy over the entire test set is 
computationally expensive, and thus we sampled $200$ random points from the 
test set (these did not necessarily have the same label, nor were they 
required to be agreement points for the ensemble). For each sample, we created 
a verification query to check the robust accuracy of the new ensembles
around the point, compared to the original ensemble. The results are
plotted in Fig.~\ref{fig:DigitMnistResultsRandomBatchEnsemble}, and
indicate that the new ensembles demonstrated \emph{significantly higher} robust 
accuracy on the tested points. These results
validate our claim that a scoring metric based on agreement points is
useful in improving the ensemble's robustness also on other,
``harder'', input points. Our analysis also indicates that the
improved robustness results originated not only from $\epsilon$-balls
around inputs labeled as ``0'', but from other labels as well. In
fact, the gain in robustness was not just in quantity, but also in
quality: for almost all cases, whenever $\ensemble_1$ proved robust
around an input, so did $\ensemble_1^{2\rightarrow 9}$ and
$\ensemble_1^{5\rightarrow 9}$. This indicates that the improved
robustness originated from inputs on which $\ensemble_1$ was prone to
err.

Next, we turned our attention to $\ensemble_2$, and computed the
uniqueness scores for each of its members (see
Table~\ref{table:DigitMnistUniqueness}).  This time we conducted a
``reverse'' experiment: we identified the two \emph{best} members of
$\ensemble_2$, i.e. the two networks that had the highest uniqueness
scores, and were consequently the least prone to err simultaneously. These turned out to be networks $N_9$ and $N_{10}$. Next, we
replaced each of these networks with each of the networks
$\{N_1,\ldots, N_5\}$, in order to identify a network that, when inserted
into $\ensemble_2$, achieved a lower score than $N_9$ and
$N_{10}$. $N_4$ turned out to be such a network.
We created the two modified ensembles,
$\ensemble_2^{9\rightarrow 4}=\{N_6,N_7,N_8,N_4,N_{10}\}$ and
$\ensemble_2^{10\rightarrow 4}=\{N_6,N_7,N_8,N_9,N_4\}$, and compared
their robust accuracy to that of $\ensemble_2$ on 200 random points
from the test set. The results, depicted
in Fig.~\ref{fig:DigitMnistResultsRandomBatchEnsemble}, indicate that the ensemble's
robust accuracy decreased significantly, as expected. 

In both aforementioned experiments, we also computed the \emph{accuracy}
(as opposed to \emph{robust accuracy}) of the new ensembles, by
evaluating them over the test set. All new ensembles had an accuracy
that was on par with that of the original ensembles --- specifically,
within a range of $\pm 0.2\%$ from the original ensembles' accuracy.

\begin{table*}[t]
  \centering
  \caption{Accuracy and uniqueness scores for the
    Fashion-MNIST networks. Uniqueness scores are measured with respect to the
  ensemble (either $\ensemble_3$ or $\ensemble_4$).}
  \scalebox{1.0}{
    \begin{tabular}{l|ccccccccccccc|}
        &
  \multicolumn{5}{c}{$\ensemble_3$} &&&&
  \multicolumn{5}{c|}{$\ensemble_4$} 
  \\
      \cline{2-6} 
      \cline{10-14}
        &
  $N_{11}$ &
  $N_{12}$ &
  $N_{13}$ &
  $N_{14}$ &
  $N_{15}$ &
        &&&
  $N_{16}$ &
  $N_{17}$ &
  $N_{18}$ &
  $N_{19}$ &
  $N_{20}$
  \\
  \midrule
  Accuracy &
  87.14\% &
  87.13\% &
  87.53\% &
  87.34\% &
  87.3\% &
      &&&
  87.05\% &
  87.32\% &
  87.35\% &
  87.34\% &
  87.11\% 
  \\
  US &
  70.63\% &
  71.5\% &
   \textbf{{\color{red}69.75\%}} &
  70.88\% &
   \textbf{{\color{green}73.25\%}} &
      &&&
  67.38\% &
  72.38\% &
  \textbf{{\color{green}80.13\%}} &
  71.38\% &
  \textbf{{\color{red}66.75\% }}
 \end{tabular}
}
\label{table:FashionMnistUniqueness}
\end{table*}








\mysubsubsection{Fashion-MNIST.}  For the second part of our
evaluation, we trained $10$ independent DNNs
$\{N_{11},\ldots,N_{20}\}$ over the Fashion-MNIST
dataset~\cite{XiRaVo17}, which includes 28$\times$28 grayscale images
of $10$ clothing categories (``Coat'', ``Dress'', etc.), and is
considered more complex than the MNIST dataset.  Each DNN had the same
architecture as the MNIST-trained DNNs, and achieved an accuracy of
$87.05\%$--$87.53\% $ (see
Table~\ref{table:FashionMnistUniqueness}). We arbitrarily constructed
two distinct ensembles, $\ensemble_3=\{N_{11},\ldots,N_{15}\}$ and
$\ensemble_4=\{N_{16},\ldots,N_{20}\}$, with an accuracy of $88.22\%$
and $88.48\%$, respectively.
 
 Next, we again computed the $\US$ values of each of the networks. The
 results, which appear in full in
 Tables~\ref{table:FashionMnistEnsembleThreeUniquenessScores} and~\ref{table:FashionMnistEnsembleFourUniquenessScores} of the Appendix, indicate a
 high variance among the uniqueness scores of the members of
 $\ensemble_4$, as compared to the relatively similar scores of
 $\ensemble_3$'s members. We thus chose to focus on $\ensemble_4$.
 Based on the computed $\US$ values, we identified  $N_{20}$ as its
 least unique DNN; and, by replacing $N_{20}$ with each of the
 five networks not currently in $\ensemble_4$, identified that
 $N_{15}$ is a good candidate for replacing $N_{20}$. Performing our
 validation step over $\ensemble_4^{20 \rightarrow 15}$ revealed that
 its robust accuracy has indeed increased. Running the ``reverse''
 experiment, in which $\ensemble_4$'s most unique member is replaced
 with a worse candidate, led us to consider the ensemble
 $\ensemble_4^{18 \rightarrow 13}$, which indeed demonstrated lower
 robust accuracy than the original ensemble. For additional details, see
 Section~\ref{sec:appendix:additionalEvaluationDetails} of the Appendix.




\begin{figure}[htp]
  \begin{center}
   \includegraphics[width=1.1\linewidth]{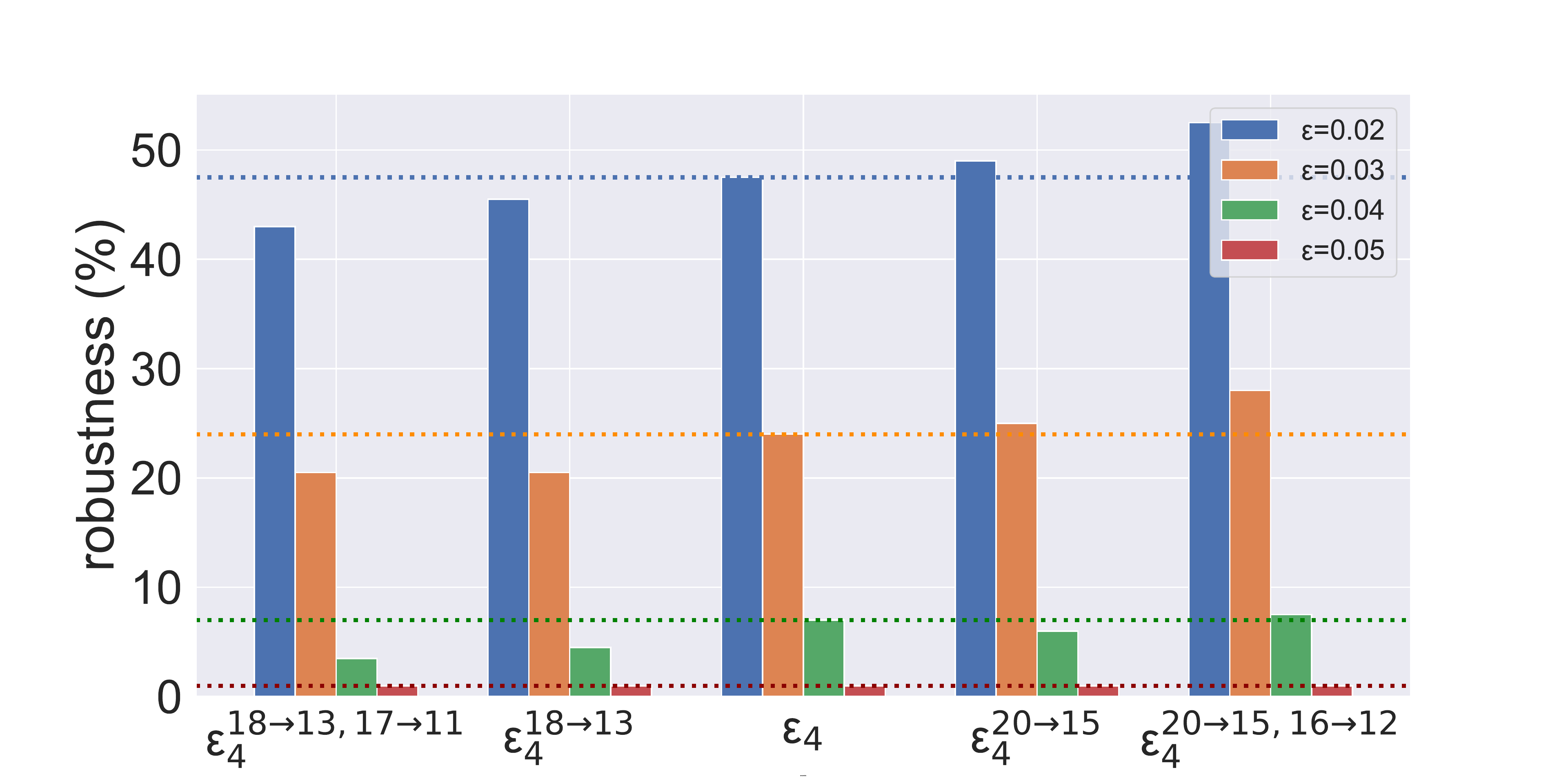}
  \end{center}
  \caption{The original ensemble $\ensemble_4$ (center),
    ensembles modified to gain robust accuracy (right),
    and ensembles modified to reduce robust accuracy (left).}
  \label{fig:FashionMnistResults}
   \end{figure}
 For the final step of our experiment, we used our approach to
 iteratively switch two members of an ensemble. Specifically, after
 creating $\ensemble_4^{{20}\rightarrow 15}$, which had higher robust
 accuracy than $\ensemble_4$, we re-computed the $\US$ scores of its
 members, and identified again the least unique member --- in this
 case, $N_{16}$. Per our computation, the best candidate for replacing
 it was $N_{12}$. The resulting ensemble, namely
 $\ensemble_4^{20\rightarrow 15, 16\rightarrow 12}$, indeed demonstrated
 higher robust accuracy than both its predecessors. Performing another
 iteration of the ``reverse'' experiment yielded ensemble
 $\ensemble_4^{18\rightarrow 13,17\rightarrow 11}$, with poorer robust
 accuracy. The results appear in
 Fig.~\ref{fig:FashionMnistResults}. We note that the only discrepancy, namely the
 robust accuracy of $\ensemble_4^{20\rightarrow 15}$ being lower than
 that of $\ensemble_4$ for $\epsilon=0.04$, is due to timeouts.

 Similarly to the MNIST case, the new ensembles in the Fashion-MNIST
 experiments obtained an accuracy that was on par with that of the
 original ensembles --- specifically, within a range of $\pm 0.17\%$
 from the original ensemble's accuracy.

\section{Comparison to Gradient-Based Attacks}
\label{sec:gradientAttacks}
Current state-of-the-art approaches for assessing a network's
robustness and robust accuracy rely on \emph{gradient-based attacks}
--- a popular class of algorithms that, like verification methods, are
capable of finding adversarial examples for a given neural network. In
this section we compare our verification-based approach to these
methods.

Gradient-based attacks generate adversarial examples by optimizing
(via various techniques) a loss metric over the network's output,
relative to its input. This allows these methods to
effectively search the local surroundings of a fixed input point for local
optima, which often constitute adversarial
inputs.
Gradient-based methods, such as the \textit{fast-gradient sign method}
(FGSM)~\cite{HuPaGoDuAb17}, \textit{projected gradient descent}
(PGD)~\cite{MaMaScTsVl17}, and others~\cite{MaDiMe20, KuGoBe16}, are in
widespread use due to their scalability and relative ease of
use. However, as we demonstrate here, they are often unsuitable in our
setting.





In order to evaluate the effectiveness of gradient-based methods for
measuring the robust accuracy of ensembles, we modified the common
FGSM~\cite{HuPaGoDuAb17} and I-FGSM~\cite{KuGoBe16} (``Iterative FGSM'') 
methods, thus extending them into
three novel attacks aimed at finding adversarial examples that can
fool multiple ensemble members simultaneously. We refer to these
attacks as \emph{Gradient Attack (G.A.) 1, 2, and 3}.  For a thorough
explanation of these attacks, as well as information about their
design and implementation, see
Section~\ref{sec:appendix:gradientAttack} of the Appendix.




Next, we used our three attacks to search for mutual errors of DNN
pairs --- i.e., adversarial examples that simultaneously affect a pair
of DNNs. Specifically, we applied the attacks on both datasets (MNSIT
and Fashion-MNIST), and searched for adversarial examples within
various $\epsilon$-balls around the same set of agreement points used
in our previous experiments. This allowed us to subsequently compute, via 
gradient attacks,
the mutual error scores of DNN pairs, and consequently, the uniqueness
scores of each constituent ensemble member. The results of the total
number of adversarial inputs found (\sat{} queries) are summarized in
Table~\ref{table:GradientAttacksMnistAndFashionMnist}. 
Each gradient attack typically took a few seconds to run.
We also
provide further details regarding the uniqueness scores computed by
the three gradient-based methods in
Section~\ref{sec:appendix:gradientAttack} of the Appendix, and in our
accompanying artifact~\cite{ArtifactRepository}.

\begin{table}[ht]
	\centering
	\caption{The number of \sat{} queries discovered when
          searching for an adversarial attack, using the three gradient attack 
          methods ($G.A.$ $1$, $2$ and $3$), and our verification approach.}
	\scalebox{1.0}{
	
		\begin{tabular}{>{\centering}p{2.5cm}| ll*{6}{c}r}
		
        Experiment   & G.A.~$1$ & G.A.~$2$ & G.A.~$3$ & verification \\
        \hline

        \textbf{}
        &  
        & 
        &
        & \\
        
        MNIST
        & 1,333  
        & 3,886
        & 5,574
        & \textbf{16,826} \\ 
        
        \textbf{}
        & 
        & 
        & 
        & \\
        
        Fashion-MNIST
        & 17,190
        & 21,245
        & 22,129
        & \textbf{33,152} \\
        

        \textbf{}
        &   
        & 
        & 
        & \\
        
        \hdashline

        \textbf{}
        &   
        & 
        & 
        & \\

        Total
        & 18,523
        & 25,131
        & 27,703
        & \textbf{49,978} \\

		\end{tabular}%

	} 
	\label{table:GradientAttacksMnistAndFashionMnist}
      \end{table}%



The results in
Table~\ref{table:GradientAttacksMnistAndFashionMnist} include a
total of $108000$ experiments, on all ensemble pairs.\footnote{
The $108000$ experiments consist of ${10 \choose 2}$ pairs, times $200$ 
agreement points, times $6$
perturbation sizes, times $2$ datasets.}
In these experiments, our verification-based  approach returned
$49978$ \sat{} results, while the strongest gradient-based method (gradient
attack number $3$) returned only $27703$ \sat{} results --- a $44\%$
decrease in the number of counterexamples found.
This discrepancy is on par with previous research~\cite{WuZeKaBa22},
which indicates that gradient-based methods may err significantly when
used for adversarial robustness analysis. This phenomenon manifests
strongly in our setting, which
involves many small and medium-sized perturbations that gradient-based
approaches struggle with~\cite{DoLiPaSuZhHuLi18}.


The reduced precision afforded by gradient-based approaches can, in
some cases, lead to sub-optimal ensemble selection choices when
compared to our verification-based approaches. Specifically, even if a
gradient-based approach produces a uniqueness score ranking that
coincides with the one produced using verification, the dramatic
decrease in the number of \sat{} queries leads to much smaller mutual
error scores, and consequently --- to uniqueness score values that are
overly optimistic, and less capable of distinguishing between poor and superior
robust accuracy results.

For example, when observing the first two arbitrary ensembles on the
MNIST dataset, $\ensemble_1$ and $\ensemble_2$, the three gradient
approaches (G.A.~$1$, $2$ and $3$) respectively assign average
uniqueness scores of $\langle 95.4\%$, $97.8\%\rangle$,
$\langle 87.5\%$, $94.5\%\rangle$ and $\langle 83.1\%$,
$92.5\%\rangle$ to the two ensembles (when averaging the $\US{}$ over
all ensemble members and all perturbations). This indicates that the
robust accuracy of the two ensembles is fairly similar (see Tables in
section~\ref{sec:appendix:gradientAttack} of the appendix).  In
contrast, when using the more sensitive, verification-based approach,
we find a substantially higher number of mutual errors (see
Table~\ref{table:GradientAttacksMnistAndFashionMnist}), and
consequently, detect a much larger gap between the uniqueness scores of
the two ensembles: $55\%$ and $77\%$.

Another example that demonstrates the increased sensitivity of our
method, when compared to gradient-based approaches, is obtained by
observing the average uniqueness score of $\ensemble_3$ and
$\ensemble_4$ on the Fashion-MNIST dataset. The strongest gradient
attack that we used assigned almost identical average uniqueness
scores to both ensembles (up to a difference of $0.01\%$), while our
approach was sensitive enough to find a $2\%$ difference between the
average $\US{}$ of the two ensembles.

Finally, we note that, unlike verification-based approaches, 
gradient attacks are incomplete, and are consequently unable to return
\unsat{}. This makes them less suitable for assessing any additional
uniqueness metrics based on robust $\epsilon$-balls.  We thus argue
that, although gradient-based methods are faster and more scalable than
verification, our results showcase the benefits of using
verification-based approaches for assessing uniqueness scores and for
ensemble selection.

\section{Related Work}
\label{sec:relatedWork}

Due to its pervasiveness, the phenomenon of adversarial inputs has
received a significant amount of attention~\cite{SzZaSuBrErGoFe13,
  GoShSz14, PaMcJhFrCeSw16, PaMcGoJhCeSw17, MoSeFaFr16, FaFoWeIdMu19,
  ZuAkGu18 }. More specifically, the machine learning community has
put a great deal of effort into measuring and improving the robustness
of networks~\cite{CiBoGrDa17, MaMaScTsVl17, CoRoKo19,
  QiMaGoKrDvFaDeStKo19, WoRiKo20, ShNaGhXuDiStDaTaGo19,
  GaUsAjGeLaLaMaLe16, HaYaYuNiXuHuTsSu18, YuHaYaNiTsSu19, LuLoWaJo19,
  ShSaZhGhStJaGo19, CaKoDaKoKaAmRe22}.  The formal methods community has also 
  been
looking into the problem, by devising scalable DNN verification,
optimization and monitoring techniques~\cite{LuScHe21, AlAvHeLu20,
  AvBlChHeKoPr19, BaShShMeSa19, PrAf20, AnPaDiCh19, SiGePuVe19,
  XiTrJo18, Eh17, AmWuBaKa21, OsBaKa22, BuTuToKoMu18,
  AsHaKrMo20, ZhShGuGuLeNa20, JaBaKa20, LoMa17, RuHuKw18, IsBaZhKa22}. To the 
  best
of our knowledge, ours is the first attempt to apply DNN verification
to the setting of DNN ensembles. We note that our approach uses a DNN
verifier strictly as a black-box backend, and so its scalability will
improve as DNN verifiers become more scalable.



Obtaining DNN specifications to be verified is a difficult problem.
While some studies have successfully applied verification to
properties formulated by domain-specific
experts~\cite{KaBaDiJuKo17Reluplex, SuKhSh19, DuChSa19, AmScKa21, 
AmCoYeMaHaFaKa22, CoYeAmFaHaKa22},
most research has been focusing on \emph{universal properties}, which
pertain to every DNN-based system; specifically, local adversarial
robustness~\cite{CaKaBaDi17, GoKaPaBa18, LyKoKoWoLiDa20, SiGePuVe19},
fairness properties~\cite{UrChWuZh20}, network
simplification~\cite{GoFeMaBaKa20} and modification~\cite{ReKa22,
  GoAdKeKa20, SoTh19, YaYaTrHoJoPo21, DoSuWaWaDa20, UsGoSuNoPa21}, and
watermark resilience~\cite{GoAdKeKa20}. 




\section{Conclusion and Future Work
}
\label{sec:conclusion}


In this case-study paper, we demonstrate a novel technique for
assessing a deep ensemble's robust accuracy through the use of DNN
verification. To mitigate the difficulty inherent to verifying large
ensembles, our approach considers pairs of networks, and
computes for each ensemble member a score that indicates its tendency to make
the same errors as other ensemble members. These scores allow us to
iteratively improve the robust accuracy of the ensemble, by replacing
weaker networks with stronger ones. Our empiric evaluation indicates
the high practical potential of our approach; and, more broadly, we
view this work as a part of the ongoing endeavor for demonstrating the
real-world usefulness of DNN verification, by identifying additional,
universal, DNN specifications.

Moving forward, we plan to tackle the natural open questions raised by
our work; specifically, how our methodology for selecting robustly
accurate ensembles can be extended beyond the current greedy search
heuristic, as well as how ensembles should be selected in the context
of other performance objectives, beyond robust accuracy. We also plan
on experimenting with multiple stopping conditions for the ensemble
member replacement process; as well as explore potential synergies
between our verification-based approach and gradient-based approaches
for computing mutual error scores. In addition, we note that we are
currently extending our approach to regression
learning ensembles and deep reinforcement learning ensembles. 
Finally, we are in the process of optimizing our approach by using 
lighter-weight, incomplete verification tools (e.g.,~\cite{SiGePuVe19, 
WuOzZeIrJuGoFoKaPaBa20, ZeWuBaKa22}), which afford better scalability, and also support
parallelization. This will hopefully allow us to handle significantly larger 
DNNs and more complex datasets.


\medskip
\noindent
\textbf{Acknowledgements.}  
We thank Haoze Wu for his 
contribution to this project.  
The first three authors were partially supported by the Israel Science 
Foundation (grant 
number 683/18). 
The first author was partially supported by the Center for 
Interdisciplinary Data Science Research at The Hebrew University of Jerusalem. 
The fourth author was partially supported by funding from Huawei.

{
\bibliographystyle{abbrv}
\bibliography{ensembles}
}

\newpage
\onecolumn
\begin{appendices}



\section{Verifying an Ensemble}
\label{sec:appendix:verifyingAnEnsemble}

\begin{figure}[htp]
	\centering
	{\includegraphics[width=0.7\linewidth]{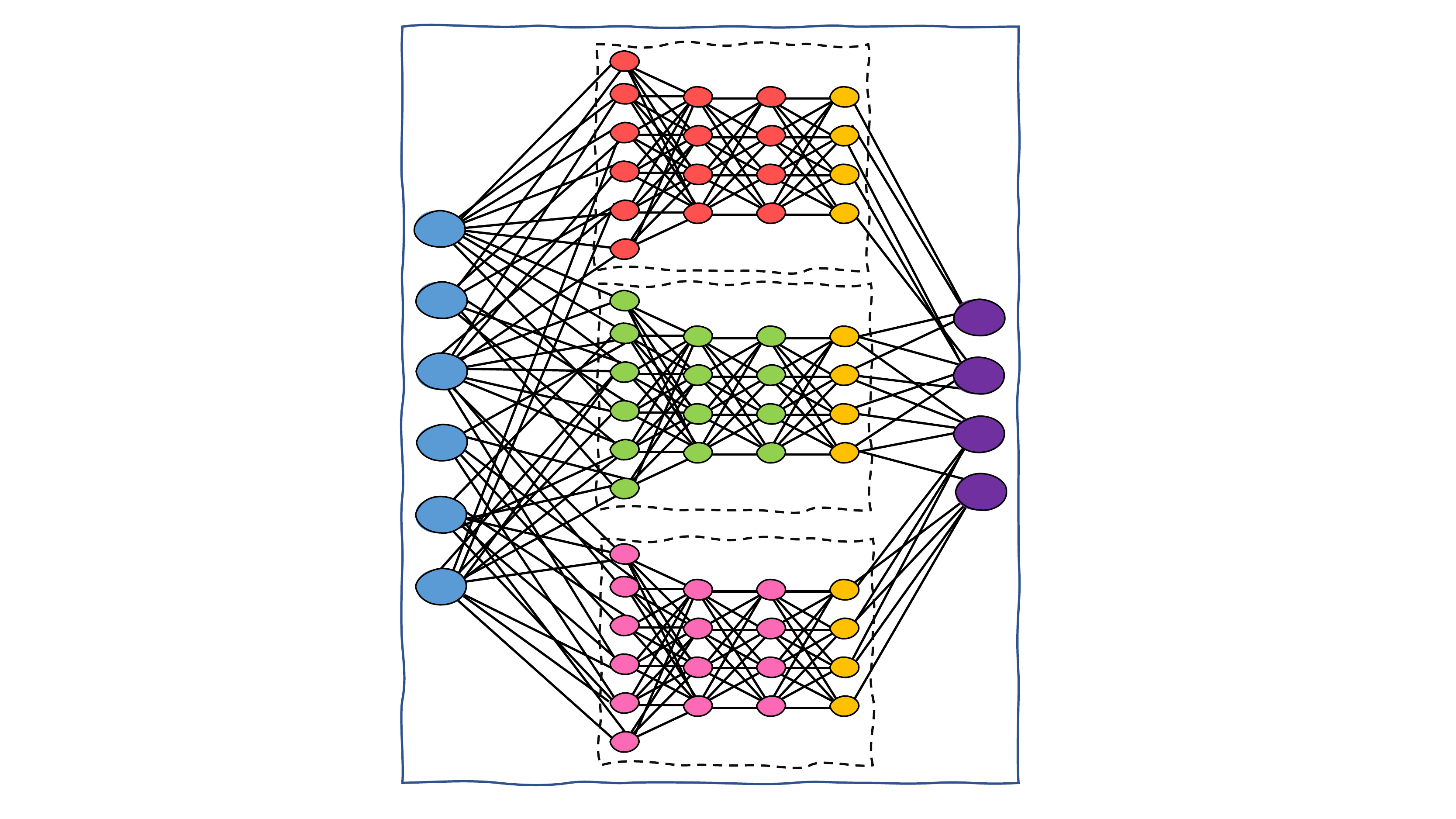}}%
	\caption{An ensemble comprised of three different DNNs, each
          depicted in a different color. The input to the ensemble is
          passed, as is, to each of the individual DNN members, and
          the output of the ensemble is the average of the outputs of
          the individual members.}%
	\label{fig:encodingAnEnsemble}%
\end{figure}

Many modern DNN verification tools receive verification queries as
$\langle P, N, Q\rangle$ triples, where $P$ is a precondition, $N$ is
the network to be verified, and $Q$ is a postcondition. In order to
verify a property of an ensemble $\ensemble=\{N_1,\ldots, N_k\}$, we
must first transform the $k$ networks into a single composite network
$N_\ensemble$, which can then be passed to the verifier. This construction is
performed as follows:
\begin{itemize}
\item By definition, all ensemble members have the same input space,
  and so their input layers all have the same dimensions. The
  composite network $N_\ensemble$ will also have an input layer of the same
  dimension.
\item Each network $N_i$ is then placed within $N_\ensemble$, with the
  composite network's input layer serving as $N_i$'s input layer. The
  $N_i$ networks do not affect each other's computation.  In
  particular, the output layer of each $N_i$ network becomes an
  internal, hidden layer of $N_\ensemble$.
  \item
    The output layer of $N_\ensemble$ is constructed to reflect the ensemble's
    aggregation mechanism. For simplicity, we focus here on the case
    where $N_\ensemble$ outputs the average of its constituent networks. In
    this case, if the networks' output domain is of dimension $t$,
    network $N_\ensemble$'s output layer is a $t$-dimensional weighted sum
    layer; and its $j$'th neuron, $n_j$, is computed as the weighted
    sum:
    \[n_j=\sum_{i=1}^k\frac{1}{k}\cdot n_j^i,\]
    where $n_j^i$ is the $j$'th output neuron of network $N_i$,
    currently encoded within $N_\ensemble$.
  \end{itemize}
  An illustration of this process appears in Fig.~\ref{fig:encodingAnEnsemble}.
  We note that similar variants of this construction have been applied
  in other contexts of DNN verification~\cite{NaKaRySaWa17, LaKa21}.

  Once $N_\ensemble$ is constructed, it can be verified for different properties, e.g., adversarial
  robustness around a given input point $x_0$, using a standard
  encoding of that verification query.

\clearpage

\section{Checking for Mutual Errors of Ensemble Members}
\label{sec:appendix:mutualErrors}

Let $N_1$ and $N_2$ be two ensemble members, and suppose we wish to
check whether these networks have a mutual error within a given
$\epsilon$-ball $B$ around point $x_0$, whose ground-truth label is
$l$. We can achieve this as follows:
\begin{itemize}
  \item We begin by constructing a composite network $N_c$ that
    effectively evaluates $N_1$ and $N_2$, side-by-side. This is
    similar to the process described in
    Section~\ref{sec:appendix:verifyingAnEnsemble}, but with two
    differences. First, this time we only compose two networks, and so
    the blowup in size is not as significant. Second, we do not
    construct an output layer that aggregates the outputs of $N_1$ and
    $N_2$; instead, the outputs of both $N_1$ and $N_2$ are concatenated into
    a single output layer of $N_c$ (which is consequently twice
    as large as the output layers of $N_1$ and $N_2$).
  \item We use the precondition $P$ to restrict the inputs of $N_c$ to
    the $\epsilon$-ball $B$. Specifically, let $x_0=\langle
    x_0^1,\ldots,x_0^n\rangle$; then:
    \[
      P = \bigwedge_{i=0}^n |x^i-x_0^i|\leq \epsilon,
    \]
    where $x=\langle x^1,\ldots,x^n\rangle$ is the input vector for $N_c$.
  \item We use the postcondition $Q$ to ensure that both networks
    $N_1$ and $N_2$ misclassify input $x$; that is, neither selects
    label $l$ as its output. This is achieved by requiring that, among
    the outputs of $N_1$, the neuron that represents $l$ is not
    assigned the maximal value; and likewise for $N_2$. More
    specifically, let $y^1_1,\ldots,y^1_r$ denote the outputs of
    $N_1$, and let $y^2_1,\ldots,y^2_r$ denote the outputs of $N_2$,
    so that the outputs of $N_c$ are
    $y^1_1,\ldots,y^1_r,y^2_1,\ldots,y^2_r$. The postcondition $Q$ in
    this case is
    \[
      \bigvee_{i\neq l}(y_i^1\geq y_l^1) \wedge
      \bigvee_{i\neq l}(y_i^2\geq y_l^2)
    \]
    Here, $y_l^1$ and $y_2^l$ represent the correct labels, and the
    postcondition requires that at least one other label be assigned a
    greater score, both in $N_1$ and in $N_2$.
 
  \end{itemize}
  .  It is
  straightforward to show that the query $\langle P,N_c,Q\rangle$ is
  \sat{} if $N_1$ and $N_2$ have a mutual error in $B$, and is
  \unsat{} otherwise. This query can be dispatched using any number of
  existing DNN verification tools.

An illustration of this process appears in Fig.~\ref{fig:mutualError}.
In this example, the precondition restricts the inputs to an $\epsilon$-ball 
around a correctly classified digit ``9''; the input is processed by the two 
networks independently; and the postcondition requires both networks to 
misclassify 
the input.

We note that in our experiments, we used the common practice of
simplifying the query's postcondition, by only considering the
disjunct $y_i\geq y_l$ for the label $i$ that achieved the
second-highest score on $x_0$ (the ``runner up''). This simplification
is solely to expedite the experiments; and the full
postcondition could be encoded, as well.

\clearpage

\section{Additional Evaluation Details}
\label{sec:appendix:additionalEvaluationDetails}

\subsection{MNIST}

\begin{figure*}[ht]%
	\centering

	{\includegraphics[scale=0.25]{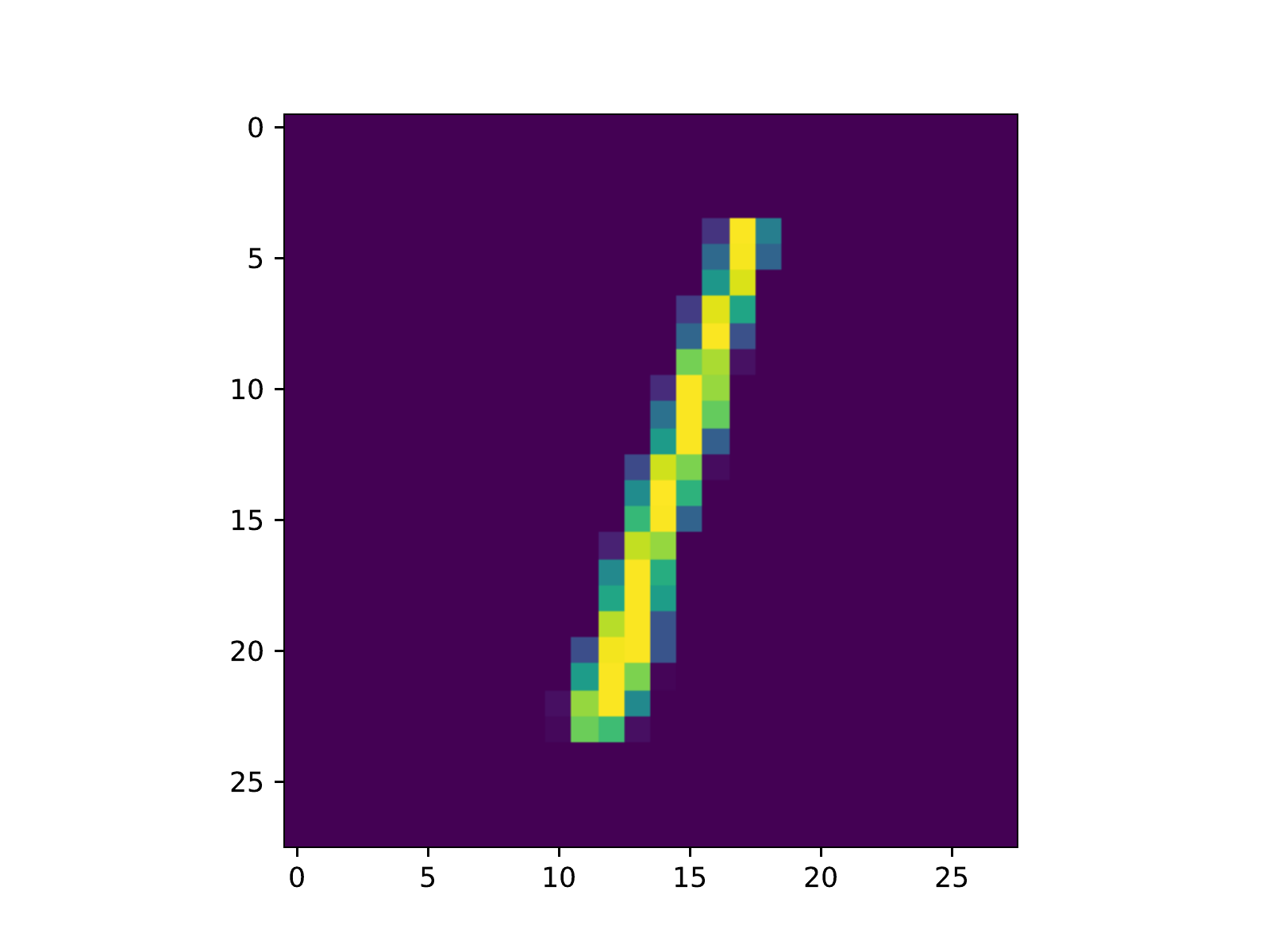}}%
	{\includegraphics[scale=0.25]{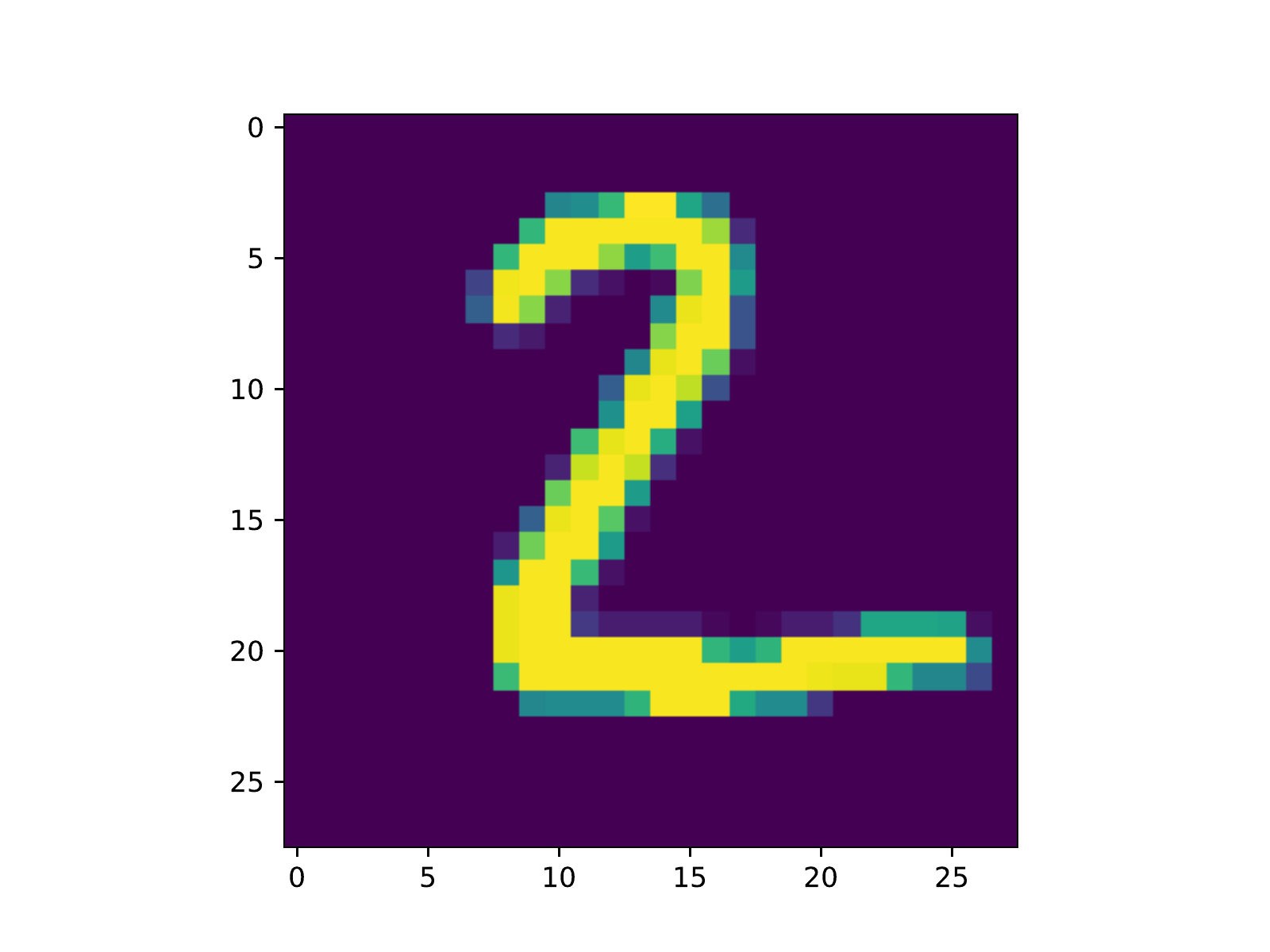}}%

	\caption{Examples of two images from the MNIST dataset. The left
          image is labeled ``1'', while the right image is labeled ``2''. All images are 28$\times$28 grayscale images.}%
	\label{fig:DigitMnistOriginal}%
\end{figure*}

\begin{figure*} [ht] %
	\centering
	
	{\includegraphics[scale=0.18]{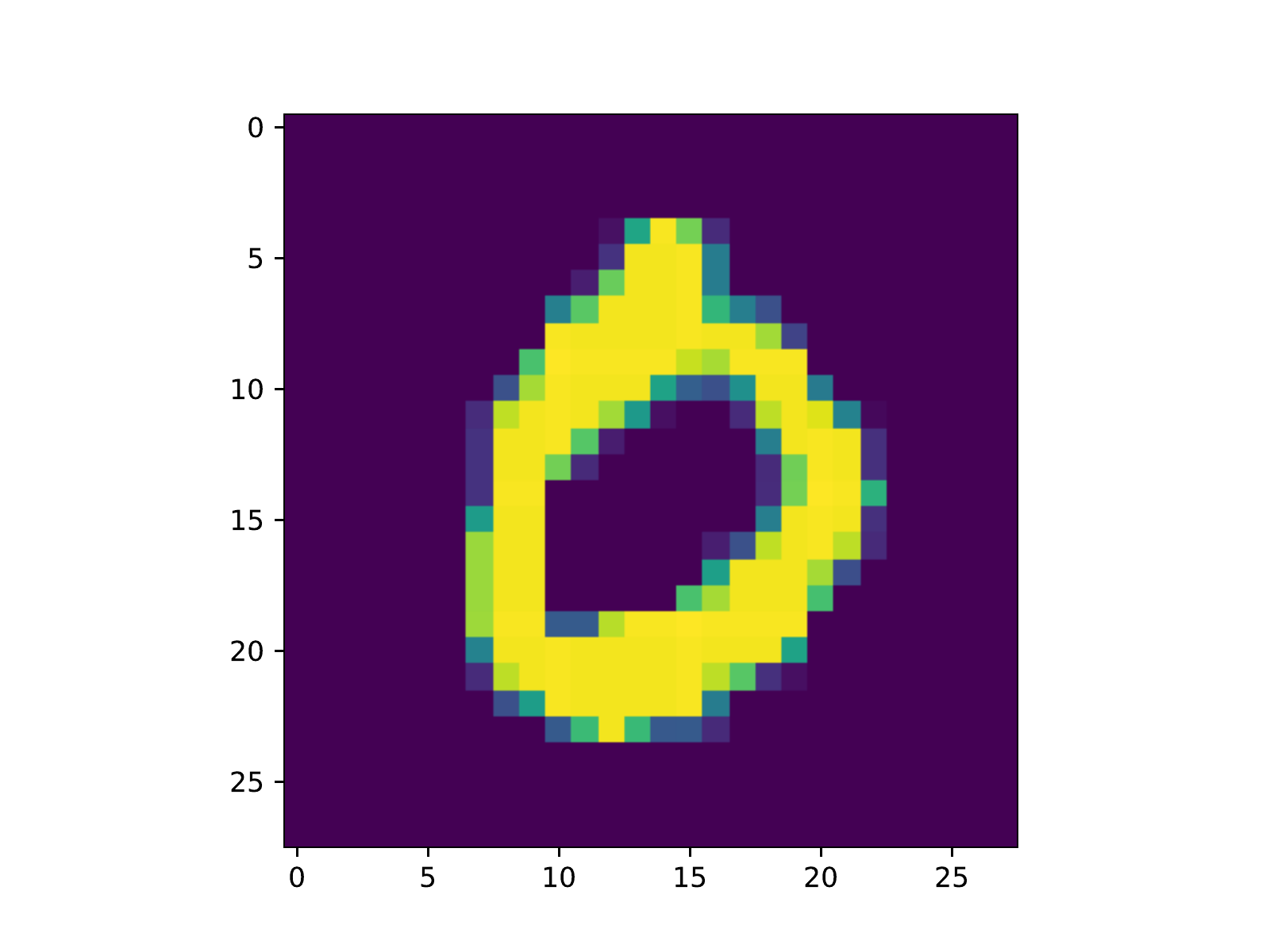}}%
	{\includegraphics[scale=0.18]{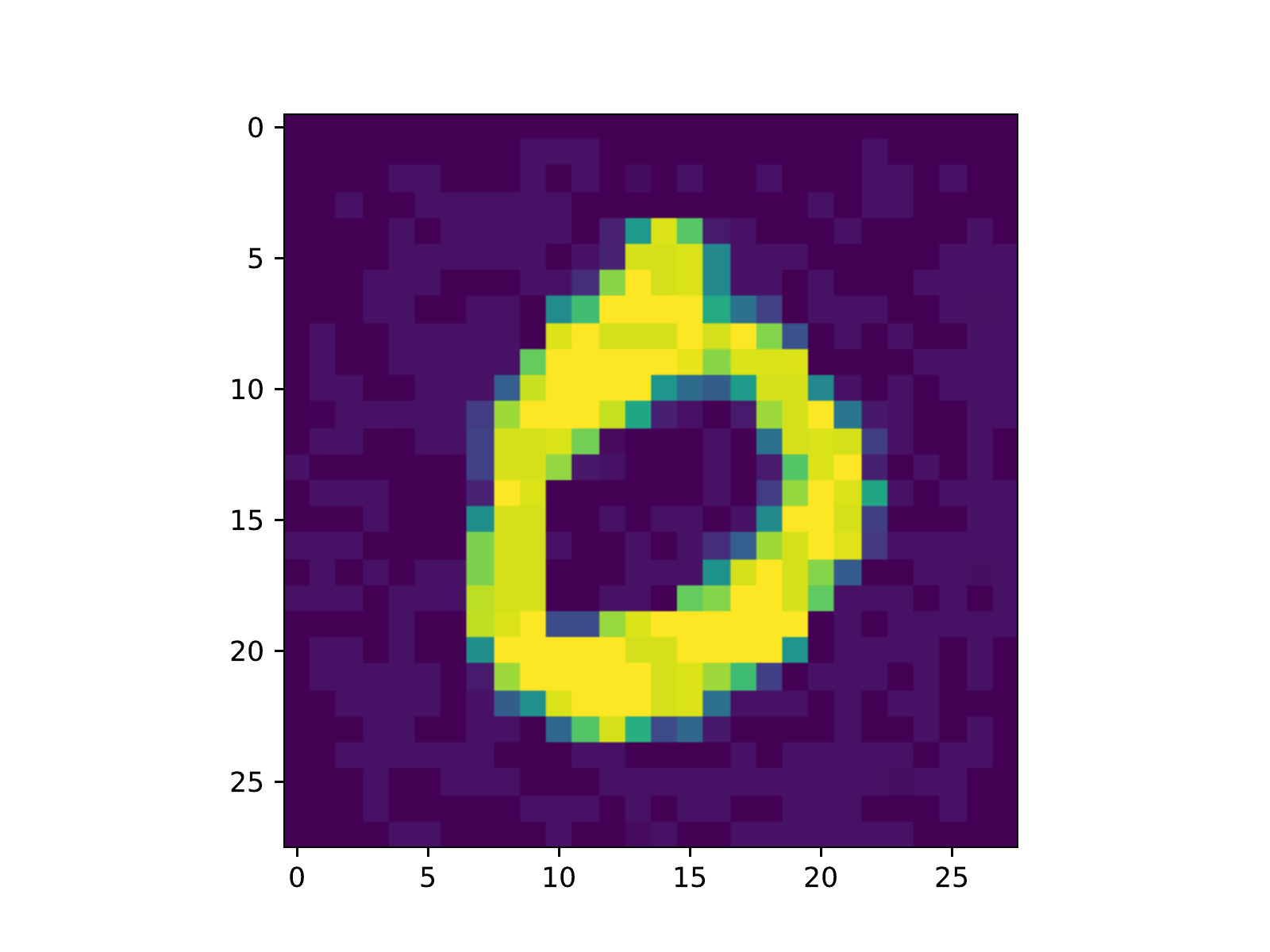}}
	{\includegraphics[scale=0.18]{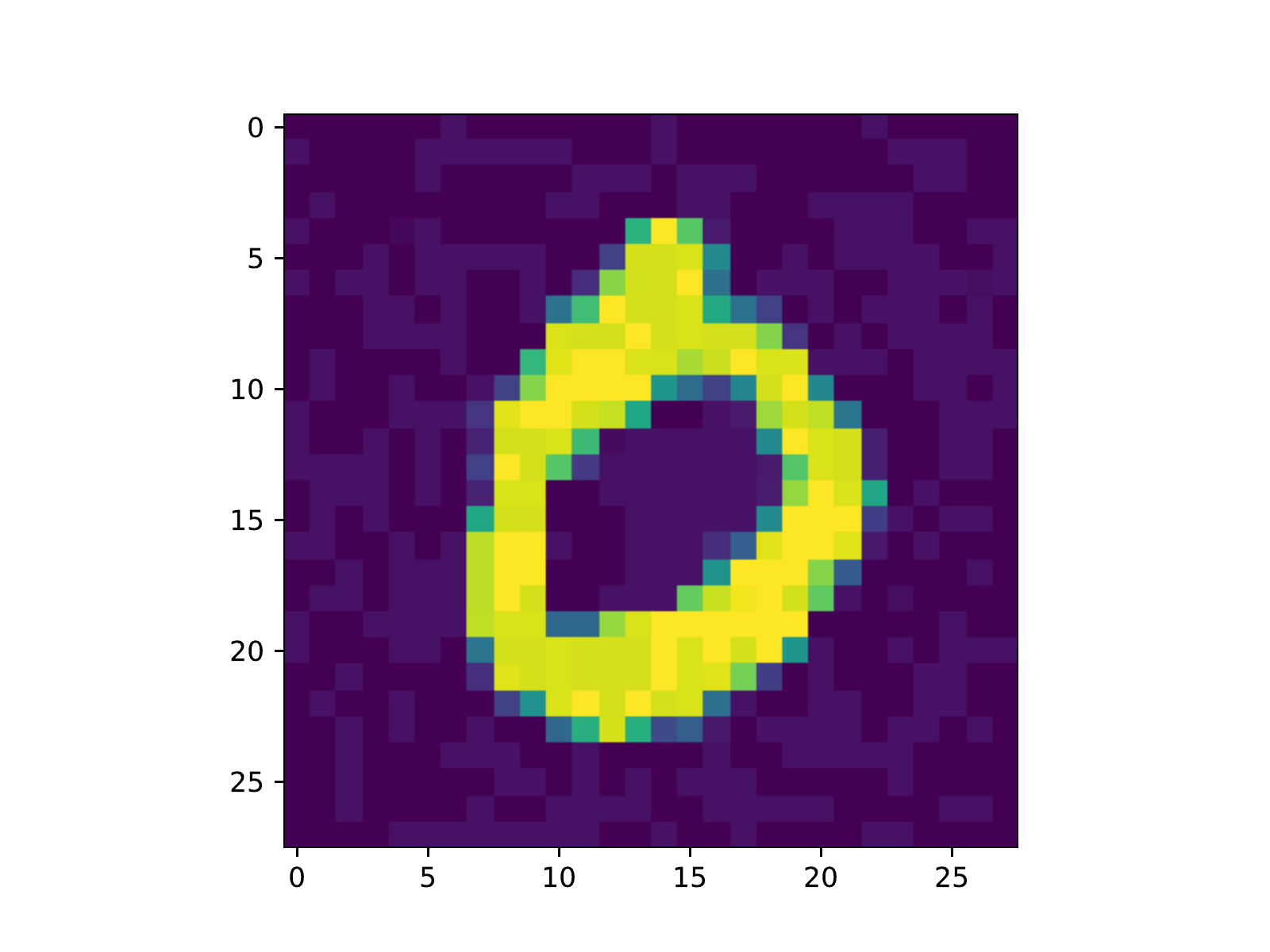}}
	{\includegraphics[scale=0.18]{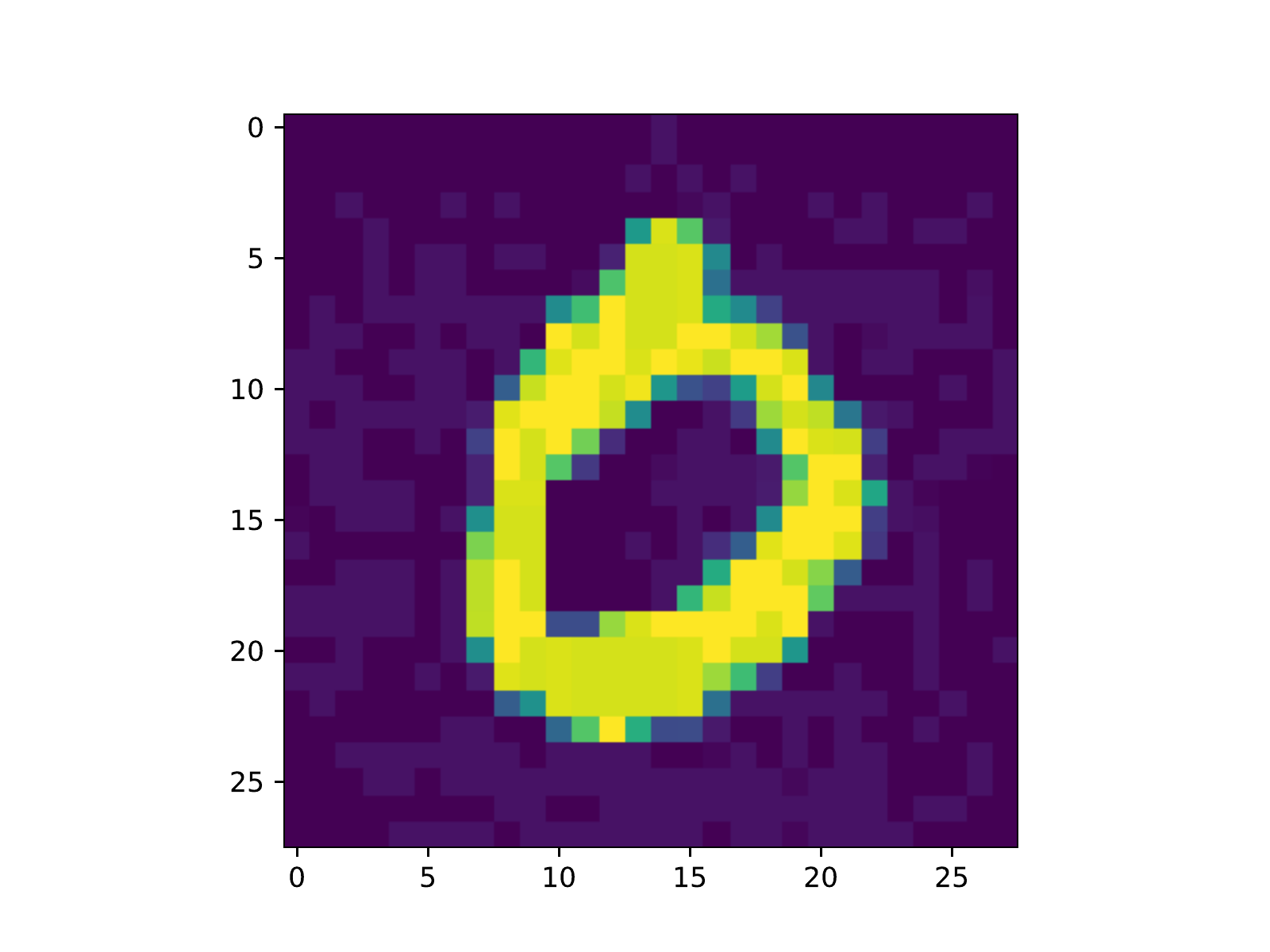}}
	
	\caption{Adversarial perturbations returned by the \sat{}
          verification queries in our experiments. From left to right
          are:
           a non-perturbed ``0'' image from the test set; and
          three $\epsilon = 0.05 $-perturbations of the original
          image, causing misclassification by members $N_1$ and
          $N_2$ (second from the left), by members $N_1$ and $N_3$
          (third from the left), and by the whole ensemble
          $\ensemble_1=\{N_1,N_2,N_3,N_4,N_5\}$ (on the right).}%
	\label{fig:DigitMnistPertubations}%
\end{figure*}

Our first set of experiments focused on the MNIST digit recognition
dataset.  Examples of inputs and perturbed inputs from this dataset
appear in Figs.~\ref{fig:DigitMnistOriginal}
and~\ref{fig:DigitMnistPertubations}.

\subsubsection{Improving the Robust Accuracy of $\ensemble_1$}
In order to improve the robust accuracy of $\ensemble_1$, we computed
the mutual error scores for each pair of ensemble members. This
computation was performed by dispatching $1200$ verification queries
for every pair, using $6$ different $\epsilon$ values. The results
appear in Table~\ref{table:DigitMnistEnsembleOneMutualErrors}, grouped
by network; e.g., the $N_1$ column shows the aggregated results of all
the pairwise queries where $N_1$ appeared. Recall that a higher number
of $\sat{}$ results (or, equivalently, a lower number of $\unsat{}$
results) indicates that a network is more prone to simultaneous errors
with its counterparts.  These results were then used to compute the
uniqueness score for each network, as presented in
Table~\ref{table:DigitMnistEnsembleOneUniquenessScores}.
Specifically, for the $200$ inputs points, per pair, per $\epsilon$
value, for each member $N_t \in \ensemble$, we calculated the
uniqueness score as:
$1- (\# \sat) \cdot \frac{1}{200 \cdot |\ensemble \setminus {N_t} |}$.
\footnote{In out experiments: $|\ensemble \setminus {N_t} |=4$, and
  the amount of all \sat{} queries for each member $N_t$ is stated,
  per $\epsilon$, in
  Table~\ref{table:DigitMnistEnsembleOneMutualErrors}.}

As the uniqueness scores show, network $N_2$ obtains the lowest score
(see Table~\ref{table:DigitMnistEnsembleOneUniquenessScores}) for each value of 
$\epsilon$, save for $\epsilon=0.01$,
where the margins are very small --- presumably because the small
perturbation size prevents any of the networks from erring, almost at
all. This clearly indicates that $N_2$ is a prime candidate for
replacement, although its original accuracy rate is actually the highest
among all networks comprising ensemble $\ensemble_1$ (see
Table~\ref{table:DigitMnistUniqueness}). Networks
$N_5$ and $N_3$ are not far behind, often obtaining the
second-lowest scores for the various epsilon values; whereas
networks $N_1$ and $N_4$ are clearly the stronger of the lot. In our
experiments we thus chose to replace $N_2$ and $N_5$. After choosing
the members to replace, we set out to search for the best replacement
candidate from $\ensemble_2$, relative to the remaining ensemble members.
As an example, we supply the
analysis at
Table~\ref{table:DigitMnistEnsembleOneReplacingN2UniquenessScores},
indicating $N_9$ is one of the two leading candidates to be added to
$\ensemble_1 \setminus \{N_2\} $ and replace $N_2$, in order to
improve the overall robust accuracy (see the left plot in
Fig.~\ref{fig:DigitMnistResultsRandomBatchEnsemble}). 
A similar
analysis to the one presented in
Table~\ref{table:DigitMnistEnsembleOneReplacingN2UniquenessScores}
can be extracted (based on our experiments summarized in the supplied 
artifact~\cite{ArtifactRepository}), and demonstrates that
$N_9$ is also one of the leading candidates to replace $N_5$ in
$\ensemble_1 \setminus \{N_5\}$.

\begin{table}[ht]
	\centering
	\caption{The results of the verification queries used to
          compute the mutual error scores of $\ensemble_1$'s
          constituent networks on the MNIST dataset.
          For each network $N_i$, the \timeout{} values are: $800-(\# \sat) - (\# \unsat)$.}
          
	\scalebox{1.0}{
		\begin{tabular}[htp]{lc|crcrcrcrcrcrcrcrcrcr}
			\toprule
			\multirow{3}{*}{\vspace*{8pt}\hspace*{9pt}$\epsilon$}
			&&& 
			\multicolumn{3}{c}{$N_1$}
			&&  
			\multicolumn{3}{c}{$N_2$}
			&&
			\multicolumn{3}{c}{$N_3$}
			&&
			\multicolumn{3}{c}{$N_4$}
			&&
			\multicolumn{3}{c}{$N_5$}
			\\
			
			\cline{4-6}
			\cline{8-10}
			\cline{12-14}
			\cline{16-18}
			\cline{20-22}
			&&&  
			\# \sat && \unsat 
			&&  
			\# \sat && \unsat 
			&&  
			\# \sat && \unsat 
			&&  
			\# \sat && \unsat 
			&&  
			\# \sat && \unsat 
			\\
			
			\midrule
			0.01
			&&&
			6 && 794
			&&
			6 && 794 
			&&
			7 && 793
			&&
			6 && 794 
			&&
			3 && 797 
			\\
			
			0.02
			&&&
			74 && 726
			&&
			93 && 707
			&&
			75 && 724
			&&
			63 && 736
			&&
			91 && 709
			\\
			
			0.03
			&&&
			270 && 517
			&&
			283 && 503
			&&
			270 && 515
			&&
			223 && 562
			&&
			258 && 523
			\\
			
			0.04
			&&&
			474 && 297  
			&&
                        507 && 266
			&&
			485 && 272
			&&
			449 && 302
			&&
			483 && 287
			\\
			
			0.05
			&&&
			621 && 142
			&&
			646 && 127
			&&
			625 && 130
			&&
			601 && 148
			&&
			623 && 139
			\\
			
			0.06
			&&&
			694 && 82
			&&
			716 && 60
			&&
			698 && 68
			&&
			692 && 79
			&&
			686 && 71
			\\
			

			\bottomrule
		\end{tabular}%
	} 
	\label{table:DigitMnistEnsembleOneMutualErrors}
\end{table}%

\begin{table}[ht]
	\centering
	\caption{The uniqueness scores for the constituent networks
          of $\ensemble_1$ on the MNIST dataset. The minimal scores are in bold.}
	\scalebox{1.0}{
		\begin{tabular}{l|l*{5}{c}r}
		
        $\epsilon$   & $N_1$ & $N_2$ & $N_3$ & $N_4$ & $N_5$ \\
        \hline
        \textbf{$0.01$}
        & 99.25  
        & 99.25 
        & \textbf{99.13}  
        & 99.25 
        & 99.63 \\ 
        
        \textbf{$0.02$}
        & 90.75  
        & \textbf{88.38} 
        & 90.63 
        & 92.13  
        & 88.63 \\ 
        
        \textbf{$0.03$}
        & 66.25  
        & \textbf{64.63} 
        & 66.25 
        & 72.13  
        & 67.75 \\
        
       \textbf{$0.04$}
        & 40.75  
        & \textbf{36.63} 
        & 39.38 
        & 43.88 
        & 39.63 \\ 
        
       \textbf{$0.05$}
        & 22.38 
        & \textbf{19.25} 
        & 21.88  
        & 24.88 
        & 22.13 \\ 
        
        \textbf{$0.06$}
        & 13.25 
        & \textbf{10.5} 
        & 12.75 
        & 13.5 
        & 14.25 \\

		\end{tabular}%
	} 
	\label{table:DigitMnistEnsembleOneUniquenessScores}
      \end{table}%

\begin{table}[ht]
	\centering
	\caption{The uniqueness scores for each replacing candidate
          from $\ensemble_2$, relative to $\ensemble_1 \setminus
          \{N_2\} $. The maximal scores are in bold.}
	\scalebox{1.0}{
	
		\begin{tabular}{>{\centering}p{2.5cm}| ll*{6}{c}r}
		
        $candidate \char`\\ 
 \epsilon$   & $0.01$ & $0.02$ & $0.03$ & $0.04$ & $0.05$ & $0.06$\\
        \hline
        \textbf{$N_6$}
        & \textbf{99.88}  
        & 94.38 
        & 85.63
        & 63.88 
        & 44.5 
        & 26.5 \\ 
        
        \textbf{$N_7$}
        & \textbf{99.88}  
        & 94.88
        & 82.63
        & 65.75
        & 44.5
        & 28.88 \\
        
        \textbf{$N_8$}
        & \textbf{99.88}
        & 96.38
        & 84.25
        & 63.88
        & 42
        & 25 \\
        
       \textbf{$N_9$}
        & \textbf{99.88}
        & \textbf{97.88}
        & \textbf{89.13}
        & 72
        & 47.88
        & 31.75 \\ 
        
        \textbf{$N_{10}$}
        & 99.63 
        & 97.25
        & 87.63 
        & \textbf{74.63} 
        & \textbf{55.25}
        & \textbf{39.25} \\

		\end{tabular}%

	} 
	\label{table:DigitMnistEnsembleOneReplacingN2UniquenessScores}
      \end{table}%

\medskip
\noindent

\subsubsection{Worsening the Robust Accuracy of $\ensemble_2$}

Next, we conducted a similar analysis of the networks comprising
$\ensemble_2$; the results appear in
Tables~\ref{table:DigitMnistEnsembleTwoMutualErrors}
and~\ref{table:DigitMnistEnsembleTwoUniquenessScores}. This time, we
set out to identify the strongest members of the ensemble. As the
maximal entries (in bold) of
Table~\ref{table:DigitMnistEnsembleTwoUniquenessScores} indicate,
networks $N_9$ and $N_{10}$ obtain higher uniqueness scores than their
counterparts. We selected $N_4$ as the replacement for $N_{10}$, as
our analysis (presented in
Table~\ref{table:DigitMnistEnsembleTwoReplacingN10UniquenessScores})
indicated that this member has a lower uniqueness score, relative to
$\ensemble_2 \setminus \{N_{10}\}$.
 We note that any of the
$\ensemble_1$ members can worsen the robust accuracy, as our metrics
indicate all members of $\ensemble_1$ achieve a lower uniqueness score
than $N_{10}$, when inserted into $\ensemble_2 \setminus
\{N_{10}\}$. 
A similar analysis to the one presented in
Table~\ref{table:DigitMnistEnsembleTwoReplacingN10UniquenessScores}
can be extracted (based on our experiments summarized in the supplied 
artifact~\cite{ArtifactRepository}), and demonstrates that
$N_4$ is also one of the leading candidates to replace $N_9$ in
$\ensemble_2 \setminus \{N_9\}$.  
As expected, the total robust accuracy
worsened when conducting the switch (see the right plot in
Fig.~\ref{fig:DigitMnistResultsRandomBatchEnsemble}).

\begin{table}[ht]
	\centering
	\caption{The results of the verification queries used to
          compute the mutual error scores of $\ensemble_2$'s
          constituent networks on the MNIST dataset.}
	\scalebox{1.0}{
		\begin{tabular}[htp]{lc|crcrcrcrcrcrcrcrcrcr}
			\toprule
			\multirow{3}{*}{\vspace*{8pt}\hspace*{9pt}$\epsilon$}
			&&& 
			\multicolumn{3}{c}{$N_6$}
			&&  
			\multicolumn{3}{c}{$N_7$}
			&&
			\multicolumn{3}{c}{$N_8$}
			&&
			\multicolumn{3}{c}{$N_9$}
			&&
			\multicolumn{3}{c}{$N_{10}$}
			\\
			
			\cline{4-6}
			\cline{8-10}
			\cline{12-14}
			\cline{16-18}
			\cline{20-22}
			&&&  
			\# \sat && \unsat 
			&&  
			\# \sat && \unsat 
			&&  
			\# \sat && \unsat 
			&&  
			\# \sat && \unsat 
			&&  
			\# \sat && \unsat 
			\\

			\midrule
			0.01
			&&&
			1 && 799
			&&
			1 && 799 
			&&
			0 && 800
			&&
			0 && 800 
			&&
			0 && 800 
			\\
			0.02
			&&&
			21 && 779
			&&
			26 && 774
			&&
			20 && 780
			&&
			9 && 791
			&&
			18 && 782
			\\
			
			0.03
			&&&
			77 && 723
			&&
			87 && 713
			&&
			78 && 721
			&&
			58 && 742
			&&
			64 && 735
			\\

			0.04
			&&&
			199 && 601 
			&&
			202 && 598
			&&
			201 && 599
			&&
			162 && 638
			&&
			166 && 634
			\\
			
			0.05
			&&&
			337 && 452
			&&
			345 && 445
			&&
			350 && 443
			&&
			327 && 460
			&&
			295 && 494
			\\
			
			0.06
			&&&
			484 && 301
			&&
			489 && 293
			&&
			503 && 288
			&&
			464 && 315
			&&
			430 && 345
			\\
			

			\bottomrule
		\end{tabular}%
	} 
	\label{table:DigitMnistEnsembleTwoMutualErrors}
\end{table}%

\begin{table}[ht]
	\centering
	\caption{The uniqueness scores for the constituent networks
          of $\ensemble_2$ on the MNIST dataset. The maximal scores are in bold.}
	\scalebox{1.0}{
		\begin{tabular}{l|l*{5}{c}r}
		
        $\epsilon$   & $N_6$ & $N_7$ & $N_8$ & $N_9$ & $N_{10}$ \\
        \hline
        $0.01$ 
        & 99.88 
        & 99.88 
        & \textbf{100}
        & \textbf{100}
        & \textbf{100}
        \\
        
        $0.02$
        & 97.38 
        & 96.75
        & 97.5
        & \textbf{98.88} 
        & 97.75 
        \\ 
        
        $0.03$     
        & 90.38 
        & 89.13 
        & 90.25 
        & \textbf{92.75}
        & 92
        \\
        
       $0.04$
        & 75.13 
        & 74.75
        & 74.88 
        & \textbf{79.75}
        & 79.25
        \\ 
        
       $0.05$      
        & 57.88 
        & 56.88 
        & 56.25  
        & 59.13 
        & \textbf{63.13} 
        \\   
        
        $0.06$
        & 39.5 
        & 38.88 
        & 37.13 
        & 42
        & \textbf{46.25}
        \\

		\end{tabular}%
	} 
	\label{table:DigitMnistEnsembleTwoUniquenessScores}
\end{table}%

\begin{table}[ht]
	\centering
	\caption{The uniqueness scores for each replacing candidate
          from $\ensemble_1$, relative to $\ensemble_2 \setminus
          \{N_{10}\} $. The minimal scores are in bold.}
	\scalebox{1.0}{
	
		\begin{tabular}{>{\centering}p{2.5cm}| ll*{6}{c}r}
		
        $candidate \char`\\ 
 \epsilon$   & $0.01$ & $0.02$ & $0.03$ & $0.04$ & $0.05$ & $0.06$\\
        \hline
        \textbf{$N_1$}
        & 100  
        & 96.38 
        & 85.25
        & 66.25 
        & 45 
        & 27.38 \\ 
        
        \textbf{$N_2$}
        & \textbf{99.75} 
        & 95.25
        & 84.75
        & \textbf{63.5}
        & \textbf{41.25}
        & \textbf{26.13} \\
        
        \textbf{$N_3$}
        & \textbf{99.75}
        & 96.38
        & 84.75
        & 65.13
        & 43.13
        & 26.75 \\
        
       \textbf{$N_4$}
        & 99.88
        & 95.75
        & 87.75
        & 69
        & 46.75
        & 29.5 \\ 
        
        \textbf{$N_5$}
        & 99.88
        & \textbf{95}
        & \textbf{83.88 }
        & 65.13
        & 44
        & 28.5 \\

		\end{tabular}%

	} 
	\label{table:DigitMnistEnsembleTwoReplacingN10UniquenessScores}
      \end{table}%

We observe that for all four novel ensembles we constructed ($\ensemble_1^{2\rightarrow 9}$, $\ensemble_1^{5\rightarrow 9}$, $\ensemble_2^{9\rightarrow 4}$, $\ensemble_2^{10\rightarrow 4}$), and presented in Fig.~\ref{fig:DigitMnistResultsRandomBatchEnsemble}, the accuracy rates range between $97.7\%$ and $98.7\%$ --- i.e., were higher than the
accuracy rates of each of the individual DNNs comprising them.
We also observe that there is a slight negative correlation between an
ensemble's robust accuracy and its accuracy: by improving the robust
accuracy of an ensemble, we risk slightly decreasing its accuracy, and vice versa. This finding is in
accordance with previous research~\cite{TsSaEnTuMa18}.

\subsection{Fashion-MNIST}


 
\begin{figure}[htp]%
	\centering

	{\includegraphics[scale=0.25]{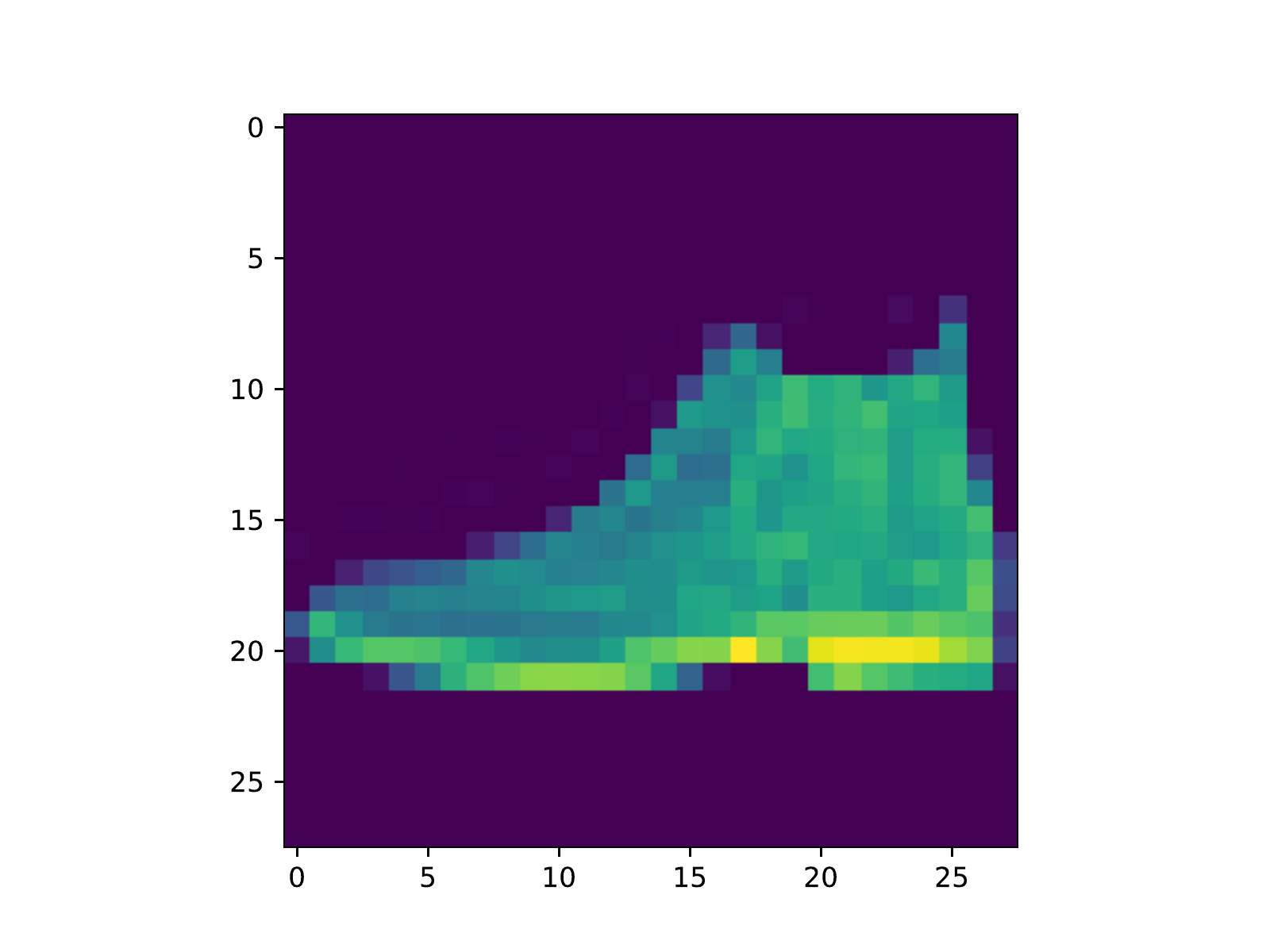}}%
	{\includegraphics[scale=0.25]{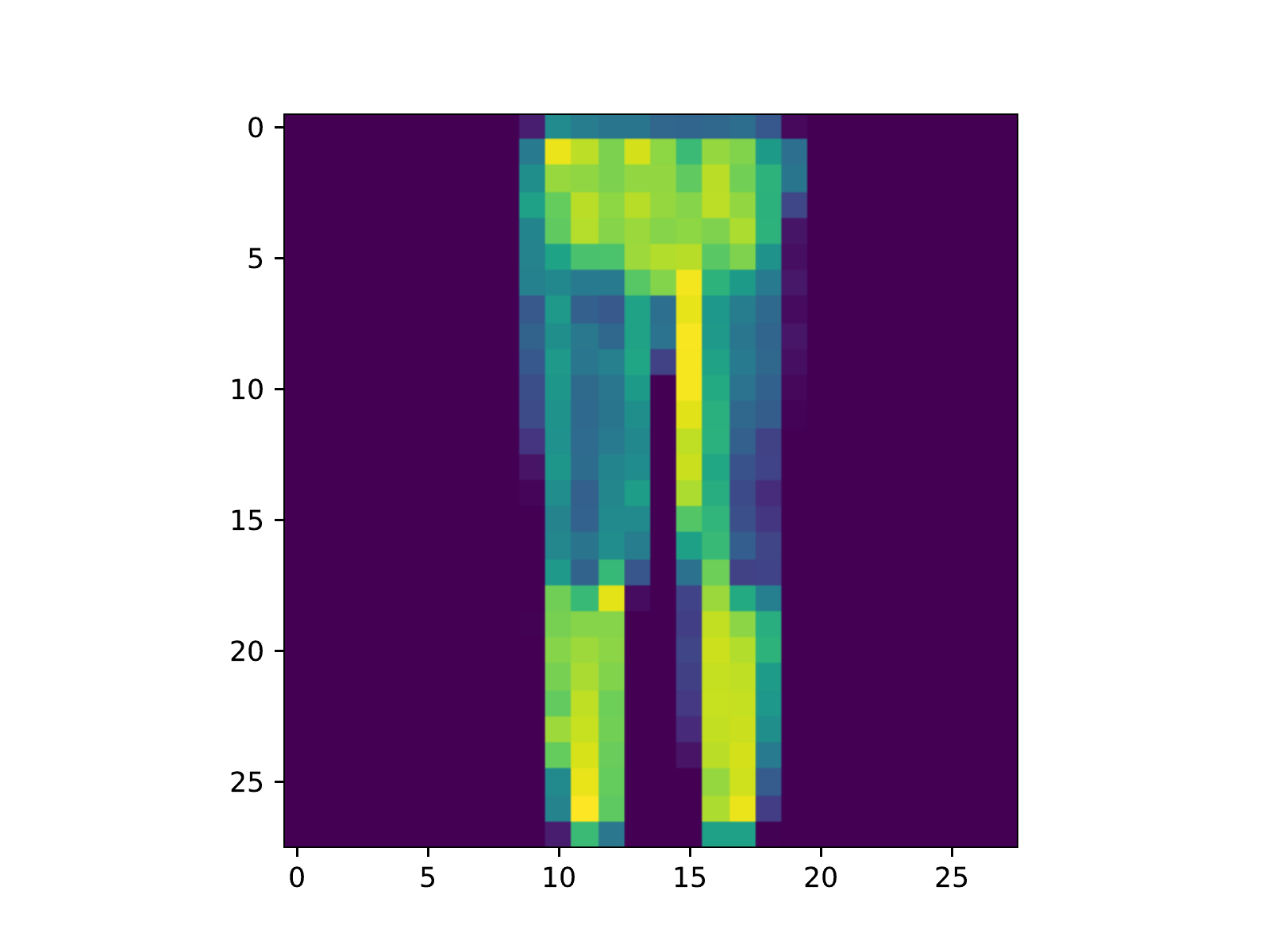}}%

	\caption{Examples of two images from the Fashion-MNIST dataset of clothing items. The left image is labeled  ``Sneaker'', while the right image is labeled ``Trouser''. All images are 28$\times$28 grayscale images.}%
	\label{fig:FashionMnistOriginal}%
\end{figure}

\begin{figure}[htp]%
	\centering
	
	{\includegraphics[scale=0.18]{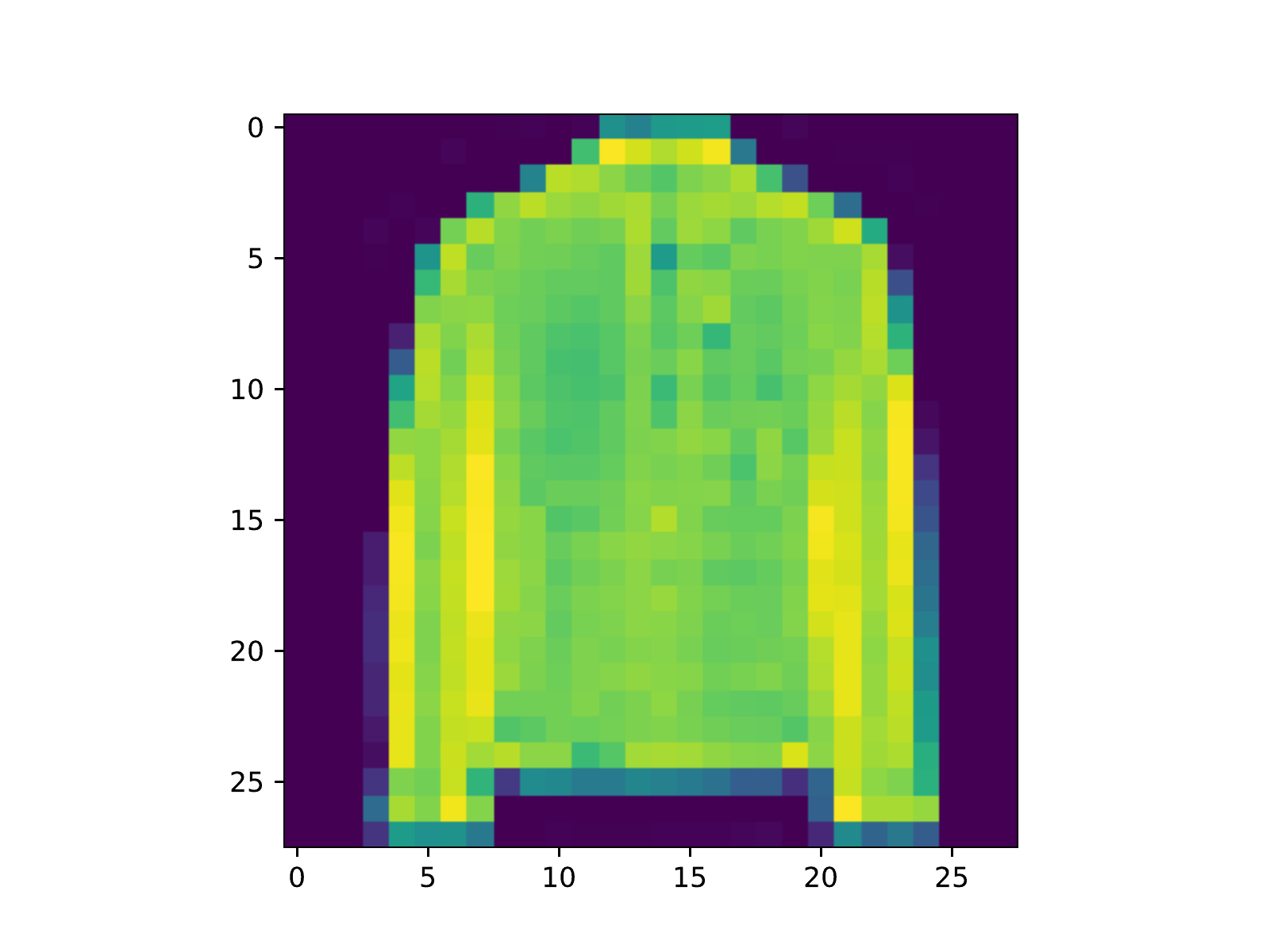}}%
	{\includegraphics[scale=0.18]{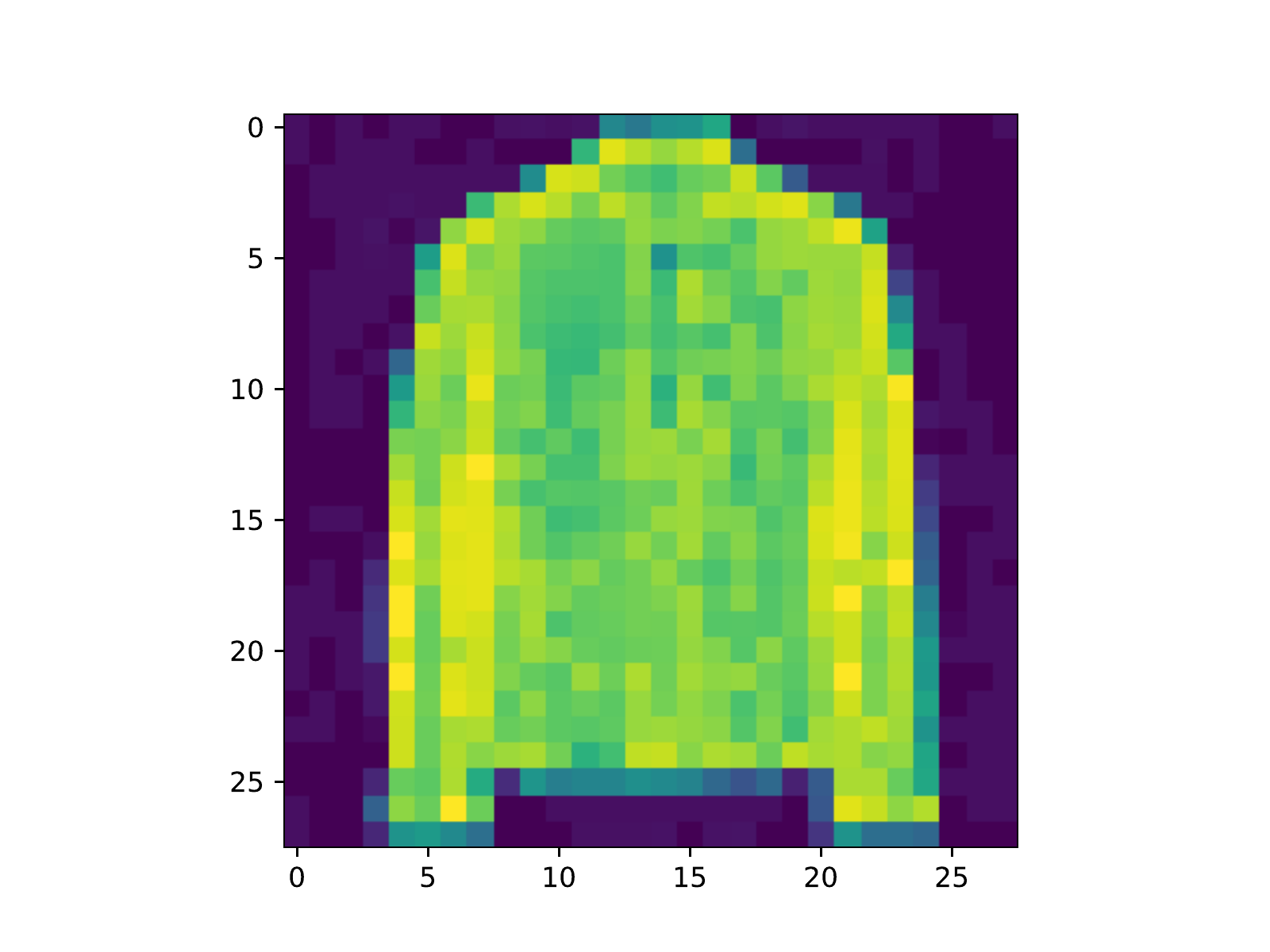}}
	{\includegraphics[scale=0.18]{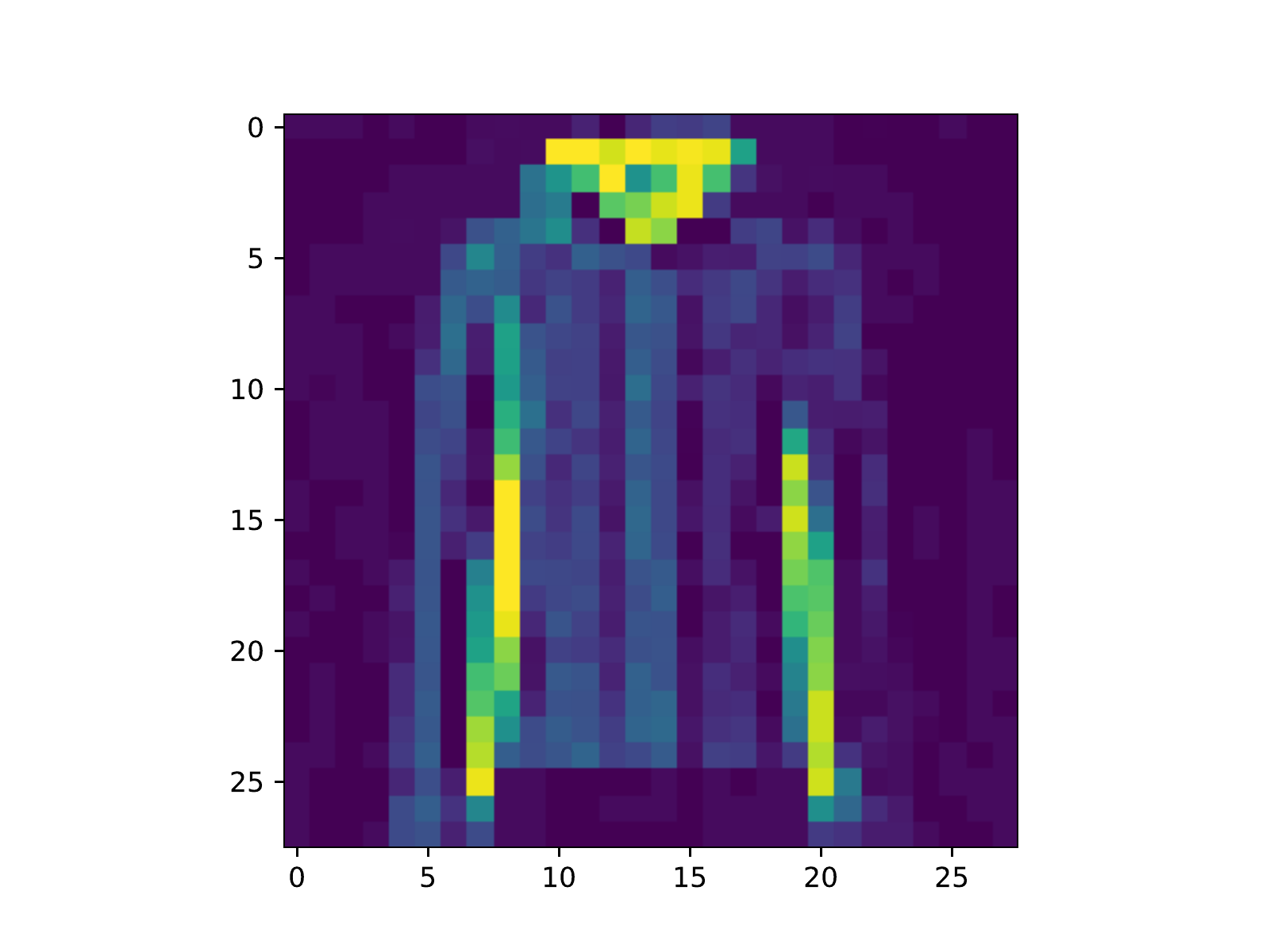}}
	{\includegraphics[scale=0.18]{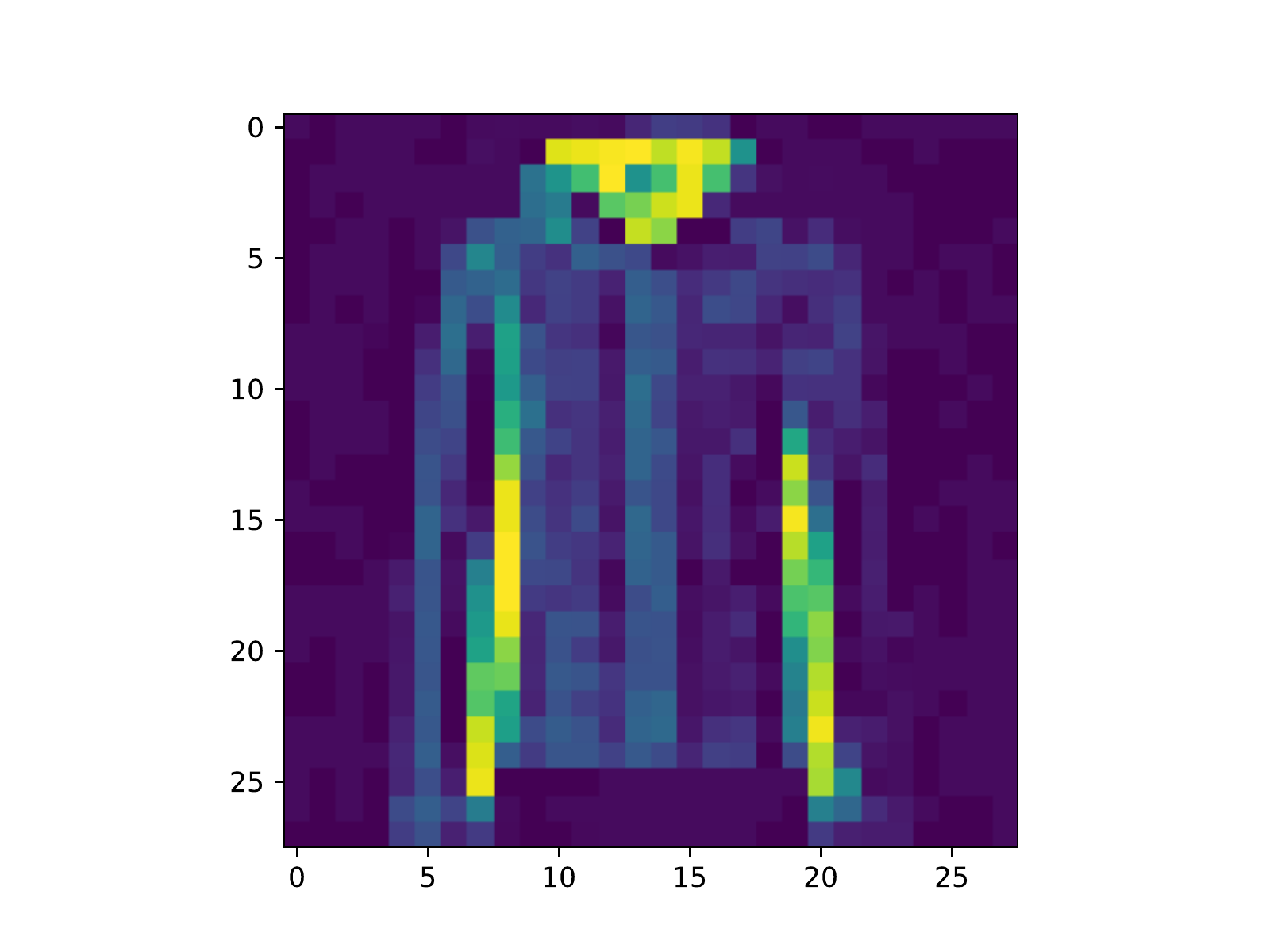}}
	
	\caption{An image
          from the Fashion-MNIST dataset labeled as a ``Coat'' (first
          on the left), and an $\epsilon = 0.04$-perturbation of that
          image misclassified by $\ensemble_3$ (second on the
          left). The two images on the right are two adversarial
          examples of another ``Coat''-labeled image, which cause a
          joint error for the pairs ($N_{13}$, $N_{14}$) and
          ($N_{14}$, $N_{15}$).}%
	\label{fig:FashionMnistPertubations}%
\end{figure}

As mentioned in Section~\ref{sec:caseStudy}, we repeated the process
on additional networks, trained on the Fashion-MNIST dataset (examples
of inputs and perturbed inputs from this dataset appear in
Figs.~\ref{fig:FashionMnistOriginal} and~\ref{fig:FashionMnistPertubations}).
In this experiment, we noticed a low variance among the relative
uniqueness scores of the members comprising $\ensemble_3$, compared to
a larger variance among the relative uniqueness scores of the members
comprising $\ensemble_4$.  This led us to focus on $\ensemble_4$, in
order to check whether our method allows improving (or worsening) the
ensemble's robust accuracy.  The high variance among the uniqueness scores
of $\ensemble_4$'s members can be seen in
Table~\ref{table:FashionMnistEnsembleFourUniquenessScores}.

\begin{table}[ht]
	\centering
	\caption{The results of the verification queries used to
          compute the mutual error scores of $\ensemble_3$'s
          constituent networks on the Fashion-MNIST dataset.}
	\scalebox{1.0}{
		\begin{tabular}[htp]{lc|crcrcrcrcrcrcrcrcrcr}
			\toprule
			\multirow{3}{*}{\vspace*{8pt}\hspace*{9pt}$\epsilon$}
			&&& 
			\multicolumn{3}{c}{$N_{11}$}
			&&  
			\multicolumn{3}{c}{$N_{12}$}
			&&
			\multicolumn{3}{c}{$N_{13}$}
			&&
			\multicolumn{3}{c}{$N_{14}$}
			&&
			\multicolumn{3}{c}{$N_{15}$}
			\\
			
			\cline{4-6}
			\cline{8-10}
			\cline{12-14}
			\cline{16-18}
			\cline{20-22}
			&&&  
			\# \sat && \unsat 
			&&  
			\# \sat && \unsat 
			&&  
			\# \sat && \unsat 
			&&  
			\# \sat && \unsat 
			&&  
			\# \sat && \unsat 
			\\
			
			\midrule
			0.01
			&&&
			43 && 757
			&&
			40 && 760
			&&
			48 && 752
			&&
			43 && 757
			&&
			28 && 772
			\\
			
			0.02
			&&&
			235 && 565
			&&
			228 && 572
			&&
			242 && 558
			&&
			233 && 567
			&&
			214 && 586
			\\
			
			0.03
			&&&
			493 && 307
			&&
			437 && 363
			&&
			484 && 316
			&&
			469 && 331
			&&
			435 && 365
			\\
			
			0.04
			&&&
			651 && 148
			&&
            611 && 188
			&&
			655 && 145
			&&
			629 && 171
			&&
			610 && 190
			\\
			
			0.05
			&&&
			759 && 38
			&&
			720 && 75
			&&
			758 && 40
			&&
			745 && 53
			&&
			736 && 58
			\\
			
			0.06
			&&&
			795 && 3
			&&
			789 && 8
			&&
			793 && 4
			&&
			785 && 8
			&&
			788 && 7
			\\

			\bottomrule
		\end{tabular}%
	} 
	\label{table:FashionMnistEnsembleThreeMutualErrors}
\end{table}%

\begin{table}[ht]
	\centering
	\caption{The uniqueness scores for the constituent networks
          of $\ensemble_3$ on the Fashion-MNIST dataset.}
	\scalebox{1.0}{
		\begin{tabular}{l|l*{5}{c}r}
		
        $\epsilon$   & $N_{11}$ & $N_{12}$ & $N_{13}$ & $N_{14}$ & $N_{15}$ \\
        \hline
        $0.01$
        & 94.63 
        & 95
        & 94
        & 94.63 
        & 96.5
        \\
        
        $0.02$
        & 70.63 
        & 71.5
        & 69.75
        & 70.88 
        & 73.25
        \\ 
        
        $0.03$
        & 38.38 
        & 45.38 
        & 39.5
        & 41.38 
        & 45.63 
        \\
        
       $0.04$
        & 18.63 
        & 23.63 
        & 18.13 
        & 21.38 
        & 23.75
        \\ 
        
       $0.05$
        & 5.13 
        & 10
        & 5.25
        & 6.88 
        & 8
        \\
        
        $0.06$
        & 0.63 
        & 1.38 
        & 0.88 
        & 1.88 
        & 1.5
        \\

		\end{tabular}%
	} 
	\label{table:FashionMnistEnsembleThreeUniquenessScores}
      \end{table}%

The mutual error scores of $\ensemble_3$'s members appear in
Table~\ref{table:FashionMnistEnsembleThreeMutualErrors}, and
these give rise to the uniqueness scores displayed in
Table~\ref{table:FashionMnistEnsembleThreeUniquenessScores}.
For $\ensemble_4$, the mutual error scores appear in
Table~\ref{table:FashionMnistEnsembleFourMutualErrors}, and the
uniqueness scores appear in
Table~\ref{table:FashionMnistEnsembleFourUniquenessScores}.
As can be seen, member $N_{18}$ is the most unique member of
$\ensemble_4$, and replacing it with $N_{13}$, a member from
$\ensemble_3$ with a low uniqueness score (see
Table~\ref{table:FashionMnistEnsembleFourReplacingN18UniquenessScores}),
worsens the robust accuracy. In the opposite direction, for various 
$\epsilon$-sized perturbations, $N_{20}$ is
the least unique member, and replacing it with $N_{15}$, a member from
$\ensemble_3$ with a higher uniqueness score (see
Table~\ref{table:FashionMnistEnsembleFourReplacingN20UniquenessScores}),
increases the robust accuracy. As in the MNIST experiments, the
uniqueness scores of the replacing members are always compared to the
remaining members. The robust accuracy changes are presented in Fig.~\ref{fig:FashionMnistResults}.



Our results for Fashion-MNISNT suggest that an ensemble that attains
high robust accuracy on different agreement points with \emph{the
  same} label (``Coat'', in our experiments) also achieves high
accuracy on points from the test data for which this is the correct
label (even if these are not agreement points). We note, however, that
our results also indicate that, unlike in the MNIST case, robust
accuracy with respect to data points with one label does not imply
robust accuracy with respect to points with \emph{a different}
label. This is not surprising; the classification of the Fashion-MNISNT
dataset is more challenging, and so there is no reason to expect that
two members of an ensemble that rarely err together on the region of
the input that corresponds to a certain label, also seldom err
simultaneously on other regions. Our results suggest that to attain
high robust accuracy with respect to the underlying distribution, the
choice of the ensemble should be done on a validation set that consists of
agreement points whose labels are distributed similarly to the
empirically observed distribution in the training data (which is
expected to approximate the underlying distribution).


\begin{table}[ht]
	\centering
	\caption{The results of the verification queries used to
          compute the mutual error scores of $\ensemble_4$'s
          constituent networks on the Fashion-MNIST dataset.}
	\scalebox{1.0}{
		\begin{tabular}[htp]{lc|crcrcrcrcrcrcrcrcrcr}
			\toprule
			\multirow{3}{*}{\vspace*{8pt}\hspace*{9pt}$\epsilon$}
			&&& 
			\multicolumn{3}{c}{$N_{16}$}
			&&  
			\multicolumn{3}{c}{$N_{17}$}
			&&
			\multicolumn{3}{c}{$N_{18}$}
			&&
			\multicolumn{3}{c}{$N_{19}$}
			&&
			\multicolumn{3}{c}{$N_{20}$}
			\\
			
			\cline{4-6}
			\cline{8-10}
			\cline{12-14}
			\cline{16-18}
			\cline{20-22}
			&&&  
			\# \sat && \unsat 
			&&  
			\# \sat && \unsat 
			&&  
			\# \sat && \unsat 
			&&  
			\# \sat && \unsat 
			&&  
			\# \sat && \unsat 
			\\
			
			\midrule
			0.01
			&&&
			75 && 725
			&&
			48 && 752
			&&
			30 && 770
			&&
			45 && 755 
			&&
			72 && 728
			\\
			
			0.02
			&&&
			261 && 539
			&&
			221 && 579
			&&
			159 && 641
			&&
			229 && 571
			&&
			266 && 534
			\\
			
			0.03
			&&&
			529 && 271
			&&
			494 && 306
			&&
			441 && 359
			&&
			491 && 309
			&&
			529 && 271
			\\
			
			0.04
			&&&
			692 && 108
			&&
            672 && 128
			&&
			633 && 167
			&&
			664 && 136
			&&
			685 && 115
			\\
			
			0.05
			&&&
			774 && 26
			&&
			753 && 47
			&&
			739 && 59
			&&
			756 && 43
			&&
			770 && 29
			\\
			
			0.06
			&&&
			793 && 5
			&&
			787 && 10
			&&
			780 && 19
			&&
			790 && 9
			&&
			792 && 7
			\\

			\bottomrule
		\end{tabular}%
	} 
	\label{table:FashionMnistEnsembleFourMutualErrors}
\end{table}%

\begin{table}[ht]
	\centering
	\caption{The uniqueness scores for the constituent networks
          of $\ensemble_4$ on the Fashion-MNIST dataset.}
	\scalebox{1.0}{
		\begin{tabular}{l|l*{5}{c}r}
		
        $\epsilon$   & $N_{16}$ & $N_{17}$ & $N_{18}$ & $N_{19}$ & $N_{20}$ \\
        \hline
        $0.01$
        & 90.63 
        & 94
        & 96.25
        & 94.38 
        & 91
        \\
        
        $0.02$
        & 67.38 
        & 72.38 
        & 80.13 
        & 71.38 
        & 66.75
        \\ 
        
        $0.03$
        & 33.88 
        & 38.25
        & 44.88 
        & 38.63 
        & 33.88 
        \\
        
       $0.04$
        & 13.5
        & 16
        & 20.88 
        & 17
        & 14.38 
        \\ 
        
       $0.05$
        & 3.25
        & 5.88 
        & 7.63 
        & 5.5
        & 3.75
        \\
        
        $0.06$
        & 0.88 
        & 1.63 
        & 2.5
        & 1.25
        & 1
        \\

		\end{tabular}%
	} 
	\label{table:FashionMnistEnsembleFourUniquenessScores}
      \end{table}%

\begin{table}[ht]
	\centering
	\caption{The uniqueness scores for each replacing candidate from $\ensemble_3$, relative to $\ensemble_4 \setminus \{N_{18}\}$.
	The minimal scores are in bold.}
	\scalebox{1.0}{
	
		\begin{tabular}{>{\centering}p{2.5cm}| ll*{6}{c}r}
		
        $candidate \char`\\ 
 \epsilon$   & $0.01$ & $0.02$ & $0.03$ & $0.04$ & $0.05$ & $0.06$\\
        \hline
        \textbf{$N_{11}$}
        & \textbf{90.75}  
        & 66.75 
        & \textbf{31.5}
        & \textbf{14.38}
        & \textbf{3.38} 
        & \textbf{0.5} \\ 
        
        \textbf{$N_{12}$}
        & 94.25
        & 69.25
        & 44.63
        & 20.5
        & 9.25
        & 2.13 \\
        
        \textbf{$N_{13}$}
        & 91.25
        & \textbf{66.5}
        & 34.88
        & 15.13
        & 3.75
        & 0.63 \\
        
       \textbf{$N_{14}$}
        & 92.75
        & 66.88
        & 36.88
        & 18.13
        & 5.75
        & 1.38 \\ 
        
        \textbf{$N_{15}$}
        & 96.25 
        & 72.88
        & 43.88 
        & 21.5
        & 6.75
        & 0.88 \\

		\end{tabular}%

	} 
	\label{table:FashionMnistEnsembleFourReplacingN18UniquenessScores}
      \end{table}%

\begin{table}[ht]
	\centering
	\caption{The uniqueness scores for each replacing candidate from $\ensemble_3$, relative to $\ensemble_4 \setminus \{N_{20}\}$.
	The maximal scores are in bold.}
	\scalebox{1.0}{
	
		\begin{tabular}{>{\centering}p{2.5cm}| ll*{6}{c}r}
		
        $candidate \char`\\ 
 \epsilon$   & $0.01$ & $0.02$ & $0.03$ & $0.04$ & $0.05$ & $0.06$\\
        \hline
        \textbf{$N_{11}$}
        & 92.63
        & 71.5 
        & 35.88
        & 16.38
        & 4.38
        & 1.13 \\ 
        
        \textbf{$N_{12}$}
        & 95.25
        & 73.63
        & \textbf{47.5}
        & 21.63
        & \textbf{10}
        & \textbf{3} \\
        
        \textbf{$N_{13}$}
        & 93
        & 70.63
        & 38.75
        & 16.5
        & 4.63
        & 1.13 \\
        
       \textbf{$N_{14}$}
        & 94.5
        & 71.38
        & 41
        & 19.75
        & 6.38
        & 1.63 \\ 
        
        \textbf{$N_{15}$}
        & \textbf{96.75} 
        & \textbf{76.25}
        & 46.5 
        & \textbf{23.25}
        & 7.88
        & 1.63 \\

		\end{tabular}%

	} 
	\label{table:FashionMnistEnsembleFourReplacingN20UniquenessScores}
      \end{table}%

\mysubsubsection{Note.} Throughout the appendices, all the uniqueness scores 
presented are normalized relative to $100$.


\clearpage
\section{Implementation of Gradient Attacks}
\label{sec:appendix:gradientAttack}


\subsection{Attacking Multiple Networks Simultaneously}
To evaluate the mutual error of two neural networks $N_{1},N_{2}$ over
a set $A$ of agreement points, we must determine, for every point
$a\in A$, whether there exists an adversarial input
$x_{0}\in B_{a,\epsilon}$ such that both $N_{1}$ and $N_{2}$
misclassify $x_{0}$.

In order to compare our verification-driven approach to gradient-based
methods, we adjusted common gradient-based approaches to suit this
task. This was needed because gradient/optimization attacks are
usually geared toward finding adversarial inputs for a \textit{single}
neural network in question; and so we had to tailor them to search for
adversarial inputs that fool \textit{multiple} neural networks
simultaneously. We thus present here a novel framework for modifying
existing optimization attacks, originally designed to find inputs that
fool a single DNN, and extend them to support simultaneous attacks on
multiple DNNs.

A naive approach for designing such an attack is to first construct a
larger ensemble network from the individual neural networks, and then
apply existing techniques.  However, because the output of an ensemble
is determined by averaging the outputs of its individual networks, the
attack might trick the ensemble by fooling a single neural network by
a large margin, instead of fooling all of the individual neural
networks simultaneously. Consequently, we propose a different approach.

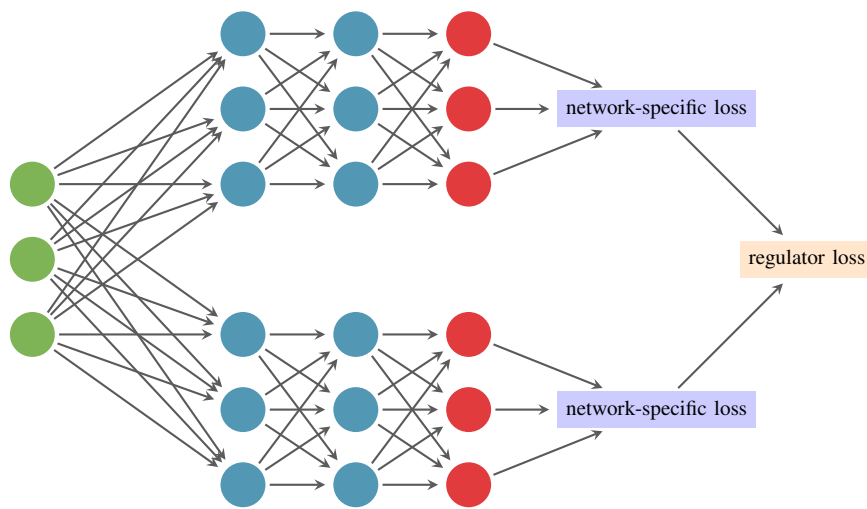
\begin{figure}[htp]
\begin{center}
	\def\WSsep{2.5cm}
	\def\relusep{1.5cm}
	\begin{tikzpicture}[shorten >=1pt,->,draw=black!50, node distance=\layersep,font=\footnotesize]
		
		\node[input neuron] (I-1) at (0,1) {};
		\node[input neuron] (I-2) at (0,0) {};
		\node[input neuron] (I-3) at (0,-1) {};
		
		\node[hidden neuron] (N1-H-1) at (\WSsep + 0.3cm,3) {};
		\node[hidden neuron] (N1-H-2) at (\WSsep + 0.3cm,2) {};
		\node[hidden neuron] (N1-H-3) at (\WSsep + 0.3cm,1) {};
		
		\node[hidden neuron] (N1-H-4) at (\WSsep + \relusep + 0.3cm,3) {};
		\node[hidden neuron] (N1-H-5) at (\WSsep + \relusep + 0.3cm,2) {};
		\node[hidden neuron] (N1-H-6) at (\WSsep + \relusep + 0.3cm,1) {};
		
		\node[output neuron] (N1-H-7) at (\WSsep + 2*\relusep + 0.3cm,3) {};
		\node[output neuron] (N1-H-8) at (\WSsep + 2*\relusep + 0.3cm,2) {};
		\node[output neuron] (N1-H-9) at (\WSsep + 2*\relusep + 0.3cm,1) {};

		\node[fill=blue!20] (N1-O-1) at (2*\WSsep + 2*\relusep + 0.3cm,2) {network-specific loss};
		
		\node[hidden neuron] (N2-H-1) at (\WSsep + 0.3cm,-1) {};
		\node[hidden neuron] (N2-H-2) at (\WSsep + 0.3cm,-2) {};
		\node[hidden neuron] (N2-H-3) at (\WSsep + 0.3cm,-3) {};
		
		\node[hidden neuron] (N2-H-4) at (\WSsep + \relusep + 0.3cm,-1) {};
		\node[hidden neuron] (N2-H-5) at (\WSsep + \relusep + 0.3cm,-2) {};
		\node[hidden neuron] (N2-H-6) at (\WSsep + \relusep + 0.3cm,-3) {};
		
		\node[output neuron] (N2-H-7) at (\WSsep + 2*\relusep + 0.3cm,-1) {};
		\node[output neuron] (N2-H-8) at (\WSsep + 2*\relusep + 0.3cm,-2) {};
		\node[output neuron] (N2-H-9) at (\WSsep + 2*\relusep + 0.3cm,-3) {};

		\node[fill=blue!20] (N2-O-1) at (2*\WSsep + 2*\relusep + 0.3cm,-2) {network-specific loss};

		\node[fill=orange!20] (L-1) at (\WSsep + 5*\relusep + 0.3cm,0) {regulator loss};
		
		\draw[nnedge] (I-1) --node[] {} (N1-H-1);
		\draw[nnedge] (I-2) --node[] {} (N1-H-1);
		\draw[nnedge] (I-3) --node[] {} (N1-H-1);
		\draw[nnedge] (I-1) --node[] {} (N1-H-2);
		\draw[nnedge] (I-2) --node[] {} (N1-H-2);
		\draw[nnedge] (I-3) --node[] {} (N1-H-2);
		\draw[nnedge] (I-1) --node[] {} (N1-H-3);
		\draw[nnedge] (I-2) --node[] {} (N1-H-3);
		\draw[nnedge] (I-3) --node[] {} (N1-H-3);
		
		\draw[nnedge] (N1-H-1) --node[] {} (N1-H-4);
		\draw[nnedge] (N1-H-2) --node[] {} (N1-H-4);
		\draw[nnedge] (N1-H-3) --node[] {} (N1-H-4);
		\draw[nnedge] (N1-H-1) --node[] {} (N1-H-5);
		\draw[nnedge] (N1-H-2) --node[] {} (N1-H-5);
		\draw[nnedge] (N1-H-3) --node[] {} (N1-H-5);
		\draw[nnedge] (N1-H-1) --node[] {} (N1-H-6);
		\draw[nnedge] (N1-H-2) --node[] {} (N1-H-6);
		\draw[nnedge] (N1-H-3) --node[] {} (N1-H-6);

		\draw[nnedge] (N1-H-4) --node[] {} (N1-H-7);
		\draw[nnedge] (N1-H-5) --node[] {} (N1-H-7);
		\draw[nnedge] (N1-H-6) --node[] {} (N1-H-7);
		\draw[nnedge] (N1-H-4) --node[] {} (N1-H-8);
		\draw[nnedge] (N1-H-5) --node[] {} (N1-H-8);
		\draw[nnedge] (N1-H-6) --node[] {} (N1-H-8);
		\draw[nnedge] (N1-H-4) --node[] {} (N1-H-9);
		\draw[nnedge] (N1-H-5) --node[] {} (N1-H-9);
		\draw[nnedge] (N1-H-6) --node[] {} (N1-H-9);

		\draw[nnedge] (N1-H-7) --node[] {} (N1-O-1);
		\draw[nnedge] (N1-H-8) --node[] {} (N1-O-1);
		\draw[nnedge] (N1-H-9) --node[] {} (N1-O-1);
		
		\draw[nnedge] (I-1) --node[] {} (N2-H-1);
		\draw[nnedge] (I-2) --node[] {} (N2-H-1);
		\draw[nnedge] (I-3) --node[] {} (N2-H-1);
		\draw[nnedge] (I-1) --node[] {} (N2-H-2);
		\draw[nnedge] (I-2) --node[] {} (N2-H-2);
		\draw[nnedge] (I-3) --node[] {} (N2-H-2);
		\draw[nnedge] (I-1) --node[] {} (N2-H-3);
		\draw[nnedge] (I-2) --node[] {} (N2-H-3);
		\draw[nnedge] (I-3) --node[] {} (N2-H-3);
		
		\draw[nnedge] (N2-H-1) --node[] {} (N2-H-4);
		\draw[nnedge] (N2-H-2) --node[] {} (N2-H-4);
		\draw[nnedge] (N2-H-3) --node[] {} (N2-H-4);
		\draw[nnedge] (N2-H-1) --node[] {} (N2-H-5);
		\draw[nnedge] (N2-H-2) --node[] {} (N2-H-5);
		\draw[nnedge] (N2-H-3) --node[] {} (N2-H-5);
		\draw[nnedge] (N2-H-1) --node[] {} (N2-H-6);
		\draw[nnedge] (N2-H-2) --node[] {} (N2-H-6);
		\draw[nnedge] (N2-H-3) --node[] {} (N2-H-6);

		\draw[nnedge] (N2-H-4) --node[] {} (N2-H-7);
		\draw[nnedge] (N2-H-5) --node[] {} (N2-H-7);
		\draw[nnedge] (N2-H-6) --node[] {} (N2-H-7);
		\draw[nnedge] (N2-H-4) --node[] {} (N2-H-8);
		\draw[nnedge] (N2-H-5) --node[] {} (N2-H-8);
		\draw[nnedge] (N2-H-6) --node[] {} (N2-H-8);
		\draw[nnedge] (N2-H-4) --node[] {} (N2-H-9);
		\draw[nnedge] (N2-H-5) --node[] {} (N2-H-9);
		\draw[nnedge] (N2-H-6) --node[] {} (N2-H-9);

		\draw[nnedge] (N2-H-7) --node[] {} (N2-O-1);
		\draw[nnedge] (N2-H-8) --node[] {} (N2-O-1);
		\draw[nnedge] (N2-H-9) --node[] {} (N2-O-1);
		
		\draw[nnedge] (N1-O-1) --node[] {} (L-1);
		\draw[nnedge] (N2-O-1) --node[] {} (L-1);
		
	\end{tikzpicture} 
\captionsetup{size=small}
\captionof{figure}{A scheme of the extended loss function for a gradient attack on multiple networks. In this case, each of the two separate networks has a predefined loss for a network-specific gradient attack. In our framework, both losses are accumulated into a novel regulating loss function which is then optimized in order to find a local optimum, which is hopefully in accordance with an adversarial attack that fools both networks simultaneously.}
\label{fig:ArchitectureOfGradientAttack}
\end{center}
\end{figure}

As a first step, our framework computes a network-specific loss for
each DNN. This is done by utilizing the loss function of the
network-specific attack that is being modified. Next, rather than
taking a step in the direction of the original loss function's
gradient, as is common in gradient-based attacks on single networks,
our method accumulates all the losses into a single ``regulating''
loss function. This function prioritizes the various single-network
losses, and performs a step in a calculated direction. Intuitively,
the regulator function should discount (but not ignore) negative
losses corresponding to networks that were already fooled, and
``focus'' on networks that are not yet fooled by the adversarial input
being constructed. An overview of this architecture (instantiated for
two networks) appears in Fig.~\ref{fig:ArchitectureOfGradientAttack}.

Given two neural networks ($N_{1}$, $N_{2}$) and an input $x$ correctly classified by both networks into class $t$:
\[N_{1}\left(x\right) = N_{2}\left(x\right) = t\], we define two search 
strategies:

\begin{itemize}
    
    \item a \textit{targeted} attack aims to cause each network to make a \textit{specific} misclassification. Formally, given two labels: $l_{1}$,$l_{2}$ such that $l_{1}\neq t$ and $l_{2} \neq t$, a targeted attack is successful if it can produce a perturbed input $x_{\epsilon}$ such that:
    \[N_{1}\left(x_{\epsilon}\right) = l_{1} \wedge N_{2}\left(x_{\epsilon}\right) = l_{2}\]

    \item  An \textit{untargeted} attack, on the other hand, aims to cause each network to make \textit{any} misclassification. Formally, an untargeted attack is successful if it can produce a perturbed input $x_{\epsilon}$ such that:
    \[N_{1}\left(x_{\epsilon}\right) \neq t \wedge N_{2}\left(x_{\epsilon}\right) \neq t\]

\end{itemize}

We note that both strategies are originally defined for optimizing attacks on a single DNN, and hence we slightly extended their definition to fit the setting in which we simultaneously optimize an attack on multiple DNNs.

In addition, we built our framework upon two popular gradient attacks:

\begin{itemize}

    \item \textit{FGSM} (Fast-Gradient Sign Method~\cite{HuPaGoDuAb17}) is a 
    highly scalable and 
    efficient attack in which the perturbation for the constructed adversarial 
    input is calculated based on moving a single step in the direction of the 
    gradient, characterizing a predefined loss on a DNN (with fixed 
    parameters). 
    
    \item \textit{I-FGSM} (Iterative FGSM~\cite{KuGoBe16}) is an iterative 
    algorithm that conducts an FGSM procedure multiple times. 

\end{itemize}

In our case, in order to search for adversarial examples that
simultaneously affect two networks around an agreement point, we
designed three novel attacks that we refer to as \textbf{Gradient
  Attack (G.A.) 1, 2, and 3}, differing from each other in the method
in which they extend FGSM or I-FGSM, and in their adversarial search
strategy (targeted or untargeted):

\begin{itemize}
    \item G.A. 1: is based on a \textit{targeted} FGSM algorithm
    \item G.A. 2: is based on a \textit{targeted} I-FGSM algorithm
    \item G.A. 3: is based on an \textit{untargeted} I-FGSM algorithm
\end{itemize}

\subsection{Tailoring and Optimizing the Loss Functions}

 The aforementioned attacks optimize the following loss functions we defined:

\begin{itemize}
    \item
    For the targeted network-specific loss function we used: 
    $l_{s}\left(N_{i}, x\right)\equiv N_{i}\left(x\right)_{t}-N_{i}\left(x\right)_{r_a}$ 
    where $t$ and $r_a$ are the indices of the true label the ``runner-up'' label (the second largest value of the output vector) of $N_{i}\left(a\right)$.
    
    \item
    For the untargeted network-specific loss function we used: 
    $l_{s}\left(N_{i}, x\right)\equiv N_{i}\left(x\right)_{t}-N_{i}\left(x\right)_{r_x}$
    where $t$ is the true label of $N_{i}\left(a\right)$ and $r_x$ is the runner-up of $N_{i}\left(x\right)$.
    
    \item 
    For the regulator loss function we used the sum of $\elu{}$ activations on the network-specific loss functions: $l_{r}\left(N_{1},N_{2},...,N_{k},x\right)=\sum^{k}_{i=1}{\elu\left(l_{s}\left(N_{i},x\right)\right)}$.
\end{itemize}

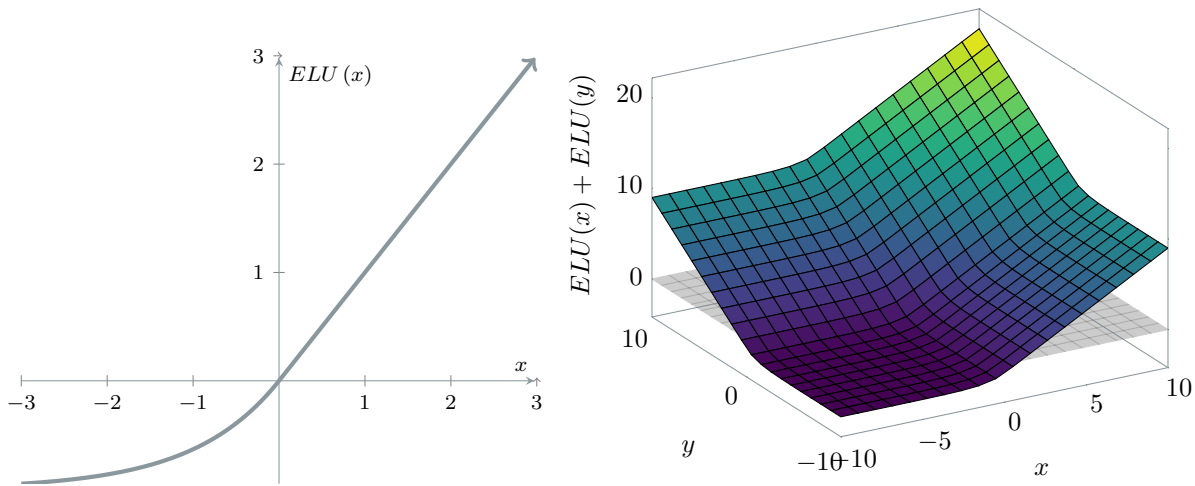
\begin{figure}[htp]
	\begin{center}
		\scalebox{1.0} {
			\def\layersep{2.0cm}
			\begin{tikzpicture}[shorten >=1pt,->,draw=black!50, node 
				distance=\layersep,font=\footnotesize]
				
				 \begin{axis}[xmin=-3,xmax=3, axis lines=middle, xlabel=$x$, ylabel=$ELU\left(x\right)$]
            \addplot[samples=100, domain=-3:3, style={ultra thick},] (x,{(x>=0) * x + (x<0) * (exp(x)-1)});
        \end{axis}
    \end{tikzpicture}
    \hfill
    \begin{tikzpicture}[declare function={g(\x)=\x<0 ? (e^\x) - 1 : \x;}, scale=1]
      \begin{axis}[shader=flat, tickwidth=0pt, view/h=-30, xlabel=$x$, ylabel=$y$, zlabel=$ELU(x)+ELU(y)$]
        \addplot3[surf,opacity=0.4, samples=20, domain=-10:10, y domain=-10:10, color=gray]{0};
        \addplot3[surf, samples=20, domain=-10:10, y domain=-10:10, colormap/viridis, draw=black]{g(x)+g(y)};
      \end{axis}
				
			\end{tikzpicture}
		}
		\captionsetup{size=small}
		\captionof{figure}{
		Left: the $ELU$ activation function.
		Right: An accumulated sum of two $ELU$ functions. Each axis corresponds to the network-specific loss attributed to one of the two networks simultaneously attacked.
		Each quadrant ($[+,-]^2$) corresponds with different regulating behavior.}
		\label{fig:EluFunctionAndMultiNetworkLoss}
	\end{center}
\end{figure}

We note that our tailored loss functions are specific compositions of the common $\elu{}$ (``Exponential Linear Unit'') activation. Formally, $\elu{}:\mathbb{R}\rightarrow\mathbb{R}$ is defined as:
\[   
\elu(x) = 
    \begin{cases} 
        x & x\geq 0 \\
        e^x-1 & x\leq 0 \\
    \end{cases}
\]

and is depicted on the left side of 
Fig.~\ref{fig:EluFunctionAndMultiNetworkLoss}. This function is often used as 
an activation function. However, we note that its properties (such as 
converging quickly) make it also an effective tool to construct our regulator 
function.

Specifically, we believe 
$l_{r}\left(N_{1},N_{2},...,N_{k},x\right)=\sum^{k}_{i=1}{\elu{}\left(l_{s}\left(N_{i},x\right)\right)}$
 to be a successful regulator function as each 
$\elu{}\left(l_{s}\left(N_{i},x\right)\right)$ term has advantageous 
properties: 

\begin{enumerate}

    \item 
    as long as its input (the network-specific loss) is positive, i.e., the specific network this loss corresponds with is not yet fooled, it outputs the loss unchanged, keeping its part in the final loss unchanged as well --- and hence ensuring that it will be optimized. 

    \item
    but, if its input is negative, that is, the network this loss corresponds 
    to is currently fooled, it will discount the final value of the regulated 
    loss function.\footnote{The discount depends on the negativity of the 
    function. 
    Intuitively, if it is close to zero, it will not discount it by much and if 
    it is very negative it will all but ignore it.}
\end{enumerate}
 
These (desirable) characteristics are depicted on the right side of 
Fig.~\ref{fig:EluFunctionAndMultiNetworkLoss}, demonstrating an accumulation of 
two network-specific losses.


\subsection{Measuring Uniqueness Scores with Gradient Attacks}

Below are tables with the main results for the uniqueness scores calculated on 
the initial ensembles of both datasets (MNIST and Fashion-MNIST).
The equivalent (verification-based) uniqueness scores of the initial MNIST ensembles ($\ensemble_1$, $\ensemble_2$) are presented in Tables~\ref{table:DigitMnistEnsembleOneUniquenessScores} and~\ref{table:DigitMnistEnsembleTwoUniquenessScores}, and the equivalent (verification-based) uniqueness scores of the initial Fashion-MNIST ensembles ($\ensemble_3$, $\ensemble_4$) are presented in Tables~\ref{table:FashionMnistEnsembleThreeUniquenessScores} and~\ref{table:FashionMnistEnsembleFourUniquenessScores} of the appendix. 

For the complete results, as well as the ensemble members' uniqueness scores after 
multiple iterations and swaps, see our supplied 
artifact~\cite{ArtifactRepository}.


\begin{table}[ht]
	\centering
		\caption{The uniqueness scores for the constituent networks
		of $\ensemble_1$ on the MNIST dataset, as calculated by \textbf{G.A. 
			1}. The average uniqueness score is \textbf{95.35}.}
	\scalebox{1.0}{
		\begin{tabular}{l|l*{5}{c}r}
			
			$\epsilon$   & $N_1$ & $N_2$ & $N_3$ & $N_4$ & $N_5$ \\
			\hline
			\textbf{$0.01$}
			& 99.88
			& 99.88
			& 100	
			& 99.75	
			& 100
			\\ 
			
			\textbf{$0.02$}
			& 99.5	
			& 98.75	
			& 99.25	
			& 99.5	
			& 99.25
			\\
			
			\textbf{$0.03$}
			& 97.75	
			& 96.25	
			& 97.25	
			& 98.88	
			& 97.63
			\\
			
			\textbf{$0.04$}
			& 95.88	
			& 92.5	
			& 94.63	
			& 96.13	
			& 94.63
			\\ 
			
			\textbf{$0.05$}
			& 92.38	
			& 88.13	
			& 91.5	
			& 94.13	
			& 92.63
			\\
			
			\textbf{$0.06$}
			& 88.88	
			& 85.13	
			& 88.5	
			& 92.5	
			& 89.5
			\\

		\end{tabular}%
	} 
	\label{table:DigitMnistEnsembleOneUniquenessScoresBasedOnGradientAttackOne}
\end{table}%

\begin{table}[ht]
	\centering
	\caption{The uniqueness scores for the constituent networks
		of $\ensemble_2$ on the MNIST dataset, as calculated by \textbf{G.A. 
			1}. 
		The average uniqueness score is \textbf{97.8}.}
	\scalebox{1.0}{
		\begin{tabular}{l|l*{5}{c}r}
			
			$\epsilon$   & $N_6$ & $N_7$ & $N_8$ & $N_9$ & $N_{10}$ \\
			\hline
			\textbf{$0.01$}
			& 100		
			& 100		
			& 100		
			& 100		
			& 100
			\\ 
			
			\textbf{$0.02$}
			& 99.75	
			& 99.88	
			& 99.75	
			& 99.88	
			& 99.75
			\\
			
			\textbf{$0.03$}
			& 99.38	
			& 98.75	
			& 98.75	
			& 99.5	
			& 98.88
			\\
			
			\textbf{$0.04$}
			& 98.38	
			& 97.5	
			& 97.88	
			& 98.88	
			& 97.88
			\\ 
			
			\textbf{$0.05$}
			& 96.63	
			& 95.75	
			& 95.75	
			& 97.25	
			& 96.63
			\\
			
			\textbf{$0.06$}
			& 93.38	
			& 92.5	
			& 92.75	
			& 94	
			& 94.63
			\\

		\end{tabular}%
	} 
	\label{table:DigitMnistEnsembleTwoUniquenessScoresBasedOnGradientAttackOne}
\end{table}%

\begin{table}[ht]
	\centering
	\caption{The uniqueness scores for the constituent networks
		of $\ensemble_3$ on the Fashion-MNIST dataset, as calculated by 
		\textbf{G.A. 
			1}. The average uniqueness score is \textbf{67.51}.}
	\scalebox{1.0}{
		\begin{tabular}{l|l*{5}{c}r}

			$\epsilon$   & $N_{11}$ & $N_{12}$ & $N_{13}$ & $N_{14}$ & $N_{15}$ 
			\\
			\hline
			\textbf{$0.01$}
			& 96.88	
			& 97.5	
			& 96.75	
			& 97	
			& 98.13
			\\ 
			
			\textbf{$0.02$}
			& 87.13	
			& 87.5	
			& 86.63	
			& 85.38	
			& 88.63
			\\
			
			\textbf{$0.03$}
			& 71.5	
			& 73.13	
			& 69.25	
			& 71.25	
			& 72.88
			\\
			
			\textbf{$0.04$}
			& 58.25	
			& 61.38	
			& 55.5	
			& 59.5	
			& 61.13
			\\ 
			
			\textbf{$0.05$}
			& 47.88	
			& 51.88	
			& 44.38	
			& 47.75	
			& 49.88
			\\
			
			\textbf{$0.06$}
			& 41.13	
			& 46.38	
			& 36.88	
			& 41.5	
			& 42.38
			\\

		\end{tabular}%
	} 
	\label{table:FashionMnistEnsembleThreeUniquenessScoresBasedOnGradientAttackOne}
\end{table}%

\begin{table}[ht]
	\centering
	\caption{The uniqueness scores for the constituent networks
		of $\ensemble_4$ on the Fashion-MNIST dataset, as calculated by 
		\textbf{G.A. 
			1}. 
		The average uniqueness score is \textbf{68.42}.}
	\scalebox{1.0}{
		\begin{tabular}{l|l*{5}{c}r}

			$\epsilon$   & $N_{16}$ & $N_{17}$ & $N_{18}$ & $N_{19}$ & $N_{20}$ 
			\\
			\hline
			\textbf{$0.01$}
			& 95	
			& 97.5	
			& 98.5	
			& 97.63	
			& 95.13
			\\ 
			
			\textbf{$0.02$}
			& 82.75	
			& 87.63	
			& 90	
			& 87.5	
			& 83.38
			\\
			
			\textbf{$0.03$}
			& 69.88	
			& 74.88	
			& 78.13	
			& 76	
			& 71.13
			\\
			
			\textbf{$0.04$}
			& 56.63	
			& 59.63	
			& 64.13	
			& 65	
			& 57.38
			\\ 
			
			\textbf{$0.05$}
			& 46.5	
			& 47.13	
			& 52.63	
			& 56.38	
			& 48.63
			\\
			
			\textbf{$0.06$}
			& 38.88	
			& 39.75	
			& 43	
			& 49.63	
			& 42.25
			\\

		\end{tabular}%
	} 
	\label{table:FashionMnistEnsembleFourUniquenessScoresBasedOnGradientAttackOne}
\end{table}%


\begin{table}[ht]
	\centering
	\caption{The uniqueness scores for the constituent networks
		of $\ensemble_1$ on the MNIST dataset, as calculated by \textbf{G.A. 
			2}. The average uniqueness score is \textbf{87.53}.}
	\scalebox{1.0}{
		\begin{tabular}{l|l*{5}{c}r}

			$\epsilon$   & $N_1$ & $N_2$ & $N_3$ & $N_4$ & $N_5$ \\
			\hline
			\textbf{$0.01$}
			& 99.88	
			& 99.88	
			& 99.75	
			& 99.5	
			& 100
			\\ 
			
			\textbf{$0.02$}
			& 98.38	
			& 97.75	
			& 98.38	
			& 98.38	
			& 98.38	
			\\
			
			\textbf{$0.03$}
			& 95	
			& 92	
			& 93.75	
			& 95	
			& 92.5	
			\\
			
			\textbf{$0.04$}
			& 87	
			& 85	
			& 87.38	
			& 89.13	
			& 86
			\\ 
			
			\textbf{$0.05$}
			& 77.75	
			& 76	
			& 78.13	
			& 81.38	
			& 77.75
			\\
			
			\textbf{$0.06$}
			& 67.5		
			& 65.13	
			& 68.25	
			& 72.38	
			& 68.5
			\\

		\end{tabular}%
	} 
	\label{table:DigitMnistEnsembleOneUniquenessScoresBasedOnGradientAttackTwo}
\end{table}%

\begin{table}[ht]
	\centering
	\caption{The uniqueness scores for the constituent networks
		of $\ensemble_2$ on the MNIST dataset, as calculated by \textbf{G.A.
			2}. 
		The average uniqueness score is \textbf{94.5}.}
	\scalebox{1.0}{
		\begin{tabular}{l|l*{5}{c}r}
			
			$\epsilon$   & $N_6$ & $N_7$ & $N_8$ & $N_9$ & $N_{10}$ \\
			\hline
			\textbf{$0.01$}
			& 99.88	
			& 99.88	
			& 100	
			& 100	
			& 100
			\\ 
			
			\textbf{$0.02$}
			& 99.5	
			& 99.75	
			& 99.38	
			& 99.75	
			& 99.38
			\\
			
			\textbf{$0.03$}
			& 98.38	
			& 97.13	
			& 98	
			& 99	
			& 98
			\\
			
			\textbf{$0.04$}
			& 95.38	
			& 93.88
			& 94.5	
			& 95.88	
			& 95.88
			\\ 
			
			\textbf{$0.05$}
			& 89.63	
			& 88.63	
			& 89.63	
			& 92.13	
			& 91.25
			\\
			
			\textbf{$0.06$}
			& 82.38	
			& 82.38	
			& 84	
			& 85.13	
			& 86.38
			\\

		\end{tabular}%
	} 
	\label{table:DigitMnistEnsembleTwoUniquenessScoresBasedOnGradientAttackTwo}
\end{table}%

\begin{table}[ht]
	\centering
	\caption{The uniqueness scores for the constituent networks
		of $\ensemble_3$ on the Fashion-MNIST dataset, as calculated by 
		\textbf{G.A. 
			2}. The average uniqueness score is \textbf{60.82}.}
	\scalebox{1.0}{
		\begin{tabular}{l|l*{5}{c}r}				
			
			$\epsilon$   & $N_{11}$ & $N_{12}$ & $N_{13}$ & $N_{14}$ & $N_{15}$ 
			\\
			\hline
			\textbf{$0.01$}
			& 96.5	
			& 97	
			& 96.25	
			& 96.38	
			& 97.63
			\\ 
			
			\textbf{$0.02$}
			& 83.63	
			& 84.38	
			& 82.63	
			& 81.75	
			& 84.88
			\\
			
			\textbf{$0.03$}
			& 64.25	
			& 65.75	
			& 61.25	
			& 65.38	
			& 66.38
			\\
			
			\textbf{$0.04$}
			& 46.25	
			& 51.75	
			& 45	
			& 50.25	
			& 50.5
			\\ 
			
			\textbf{$0.05$}
			& 37.5	
			& 42.63	
			& 35.25	
			& 39.75	
			& 39.63
			\\
			
			\textbf{$0.06$}
			& 31.25	
			& 37.13	
			& 27.88	
			& 33.38	
			& 32.38
			\\

		\end{tabular}%
	} 
	\label{table:FashionMnistEnsembleThreeUniquenessScoresBasedOnGradientAttackTwo}
\end{table}%

\begin{table}[ht]
	\centering
	\caption{The uniqueness scores for the constituent networks
		of $\ensemble_4$ on the Fashion-MNIST dataset, as calculated by 
		\textbf{G.A. 
			2}. 
		The average uniqueness score is \textbf{59.72}.}
	\scalebox{1.0}{
		\begin{tabular}{l|l*{5}{c}r}

			$\epsilon$   & $N_{16}$ & $N_{17}$ & $N_{18}$ & $N_{19}$ & $N_{20}$ 
			\\
			\hline
			\textbf{$0.01$}
			& 93.38	
			& 96.63	
			& 97.75	
			& 97	
			& 93.75
			\\ 
			
			\textbf{$0.02$}
			& 78.5	
			& 84.25	
			& 87.75	
			& 84.88	
			& 78.63
			\\
			
			\textbf{$0.03$}
			& 60.75	
			& 63.13	
			& 70.13	
			& 68	
			& 60.25
			\\
			
			\textbf{$0.04$}
			& 44.63	
			& 46.25	
			& 52.5	
			& 52.5	
			& 45.63
			\\ 
			
			\textbf{$0.05$}
			& 33.75	
			& 34.88	
			& 39.63	
			& 42.38	
			& 34.88	
			\\
			
			\textbf{$0.06$}
			& 26.75	
			& 27.63	
			& 31.25	
			& 35.13	
			& 29
			\\

		\end{tabular}%
	} 
	\label{table:FashionMnistEnsembleFourUniquenessScoresBasedOnGradientAttackTwo}
\end{table}%


\begin{table}[ht]
	\centering
	\caption{The uniqueness scores for the constituent networks
		of $\ensemble_1$ on the MNIST dataset, as calculated by \textbf{G.A. 
			3}. The average uniqueness score is \textbf{83.06}.}
	\scalebox{1.0}{
		\begin{tabular}{l|l*{5}{c}r}
			
			$\epsilon$   & $N_1$ & $N_2$ & $N_3$ & $N_4$ & $N_5$ \\
			\hline
			\textbf{$0.01$}
			& 99.63	
			& 99.63	
			& 99.5	
			& 99.5	
			& 100
			\\ 
			
			\textbf{$0.02$}
			& 97.13	
			& 95.88	
			& 97.13	
			& 97.13	
			& 96.5
			\\
			
			\textbf{$0.03$}
			& 89.75	
			& 88.13	
			& 90.25	
			& 92	
			& 89.87
			\\
			
			\textbf{$0.04$}
			& 77.88	
			& 77.75	
			& 79.38	
			& 83.5	
			& 79.75
			\\ 
			
			\textbf{$0.05$}
			& 67.88	
			& 66.25	
			& 71	
			& 74.75	
			& 70.12
			\\
			
			\textbf{$0.06$}
			& 61.88	
			& 58.38	
			& 62.13	
			& 65.88	
			& 63.25
			\\

		\end{tabular}%
	} 
	\label{table:DigitMnistEnsembleOneUniquenessScoresBasedOnGradientAttackThree}
\end{table}%

\begin{table}[ht]
	\centering
	\caption{The uniqueness scores for the constituent networks
		of $\ensemble_2$ on the MNIST dataset, as calculated by \textbf{G.A. 
			3}. 
		The average uniqueness score is \textbf{92.52}.}
	\scalebox{1.0}{
		\begin{tabular}{l|l*{5}{c}r}
				
			$\epsilon$   & $N_6$ & $N_7$ & $N_8$ & $N_9$ & $N_{10}$ \\
			\hline
			\textbf{$0.01$}
			& 99.88	
			& 99.88	
			& 100	
			& 100	
			& 100
			\\ 
			
			\textbf{$0.02$}
			& 99.38	
			& 99.38	
			& 98.75	
			& 99.75	
			& 99
			\\
			
			\textbf{$0.03$}
			& 97.5	
			& 95.88	
			& 96.88	
			& 97.88	
			& 97.63
			\\
			
			\textbf{$0.04$}
			& 93.13	
			& 91	
			& 92.25	
			& 93.63	
			& 93.5
			\\ 
			
			\textbf{$0.05$}
			& 85.38	
			& 84.88	
			& 86.38	
			& 88.63	
			& 87.25
			\\
			
			\textbf{$0.06$}
			& 77.5	
			& 78.25	
			& 79.38	
			& 80.63	
			& 82
			\\

		\end{tabular}%
	} 
	\label{table:DigitMnistEnsembleTwoUniquenessScoresBasedOnGradientAttackThree}
\end{table}%

\begin{table}[ht]
	\centering
	\caption{The uniqueness scores for the constituent networks
		of $\ensemble_3$ on the Fashion-MNIST dataset, as calculated by 
		\textbf{G.A. 
			3}. The average uniqueness score is \textbf{58.66}.}
	\scalebox{1.0}{
		\begin{tabular}{l|l*{5}{c}r}		
			
			$\epsilon$   & $N_{11}$ & $N_{12}$ & $N_{13}$ & $N_{14}$ & $N_{15}$ 
			\\
			\hline
			\textbf{$0.01$}
			& 95.5	
			& 96.25	
			& 95.25	
			& 95.75	
			& 97
			\\ 
			
			\textbf{$0.02$}
			& 81.75	
			& 82.25	
			& 81.63	
			& 79.75	
			& 82.88
			\\
			
			\textbf{$0.03$}
			& 59.75	
			& 61	
			& 59	
			& 61.88	
			& 63.13
			\\
			
			\textbf{$0.04$}
			& 42.37		
			& 48	
			& 42.38	
			& 46.5	
			& 47.25
			\\ 
			
			\textbf{$0.05$}
			& 35.62	
			& 39.88	
			& 33.38	
			& 37.75	
			& 37.88
			\\
			
			\textbf{$0.06$}
			& 30.13	
			& 35.13	
			& 27.88	
			& 31.88	
			& 31
			\\

		\end{tabular}%
	} 
	\label{table:FashionMnistEnsembleThreeUniquenessScoresBasedOnGradientAttackThree}
\end{table}%

\begin{table}[ht]
	\centering
	\caption{The uniqueness scores for the constituent networks
		of $\ensemble_4$ on the Fashion-MNIST dataset, as calculated by 
		\textbf{G.A. 
			3}. 
		The average uniqueness score is \textbf{58.67}.}
	\scalebox{1.0}{
		\begin{tabular}{l|l*{5}{c}r}
			
			$\epsilon$   & $N_{16}$ & $N_{17}$ & $N_{18}$ & $N_{19}$ & $N_{20}$ 
			\\
			\hline
			\textbf{$0.01$}
			& 92.75	
			& 96.38	
			& 97.63	
			& 96.63	
			& 93.38
			\\ 
			
			\textbf{$0.02$}
			& 76.5	
			& 83.13	
			& 86	
			& 82.25	
			& 76.38
			\\
			
			\textbf{$0.03$}
			& 58	
			& 60.25	
			& 66.25	
			& 65	
			& 57
			\\
			
			\textbf{$0.04$}
			& 41.63	
			& 44.5	
			& 51.13	
			& 50.75	
			& 43.5
			\\ 
			
			\textbf{$0.05$}
			& 33.5	
			& 35.13	
			& 38.13	
			& 41.88	
			& 35.63
			\\
			
			\textbf{$0.06$}
			& 28.25	
			& 29.25	
			& 32.38	
			& 36.38	
			& 30.5
			\\

		\end{tabular}%
	} 
	\label{table:FashionMnistEnsembleFourUniquenessScoresBasedOnGradientAttackThree}
\end{table}%

\end{appendices}

\end{document}